\documentclass[11pt]{article}

\usepackage[preprint]{acl}
\usepackage[colorinlistoftodos]{todonotes}
\usepackage{times}
\usepackage{latexsym}
\usepackage{amssymb}
\usepackage[T1]{fontenc}

\usepackage[utf8]{inputenc}
\usepackage{amsmath} 
\usepackage{microtype}
\usepackage{booktabs}

\usepackage{inconsolata}

\usepackage{graphicx}

\usepackage{svg}

\title{From Structured Records to Patient Narratives: Preserving Clinical Alignment in LLM-Generated Health Summaries}

\newcommand{\base}{\textit{Base}}
\newcommand{\constrained}{\textit{Constrained}}
\newcommand{\ragconstrained}{\textit{RAG-Constrained}}
\usepackage{placeins}

\author{
  Bayram Ayadi$^{1,2}$ \quad Florian Geissler$^{2}$ \quad Johanna Schmidhuber$^{2}$ \\
  $^1$University of Passau, Passau, Germany \\
  $^2$Fraunhofer Institute for Cognitive Systems IKS, Munich, Germany \\
  \texttt{ayadi02@ads.uni-passau.de}, \texttt{florian.geissler@iks.fraunhofer.de}
}

\begin{document}
\maketitle
\begin{abstract}

As wearable and digital health systems increasingly generate longitudinal physiological records, communicating these observations in patient-accessible language remains a significant challenge. Large Language Models (LLMs) offer a promising mechanism for translating structured clinical information into readable narratives, but this process may introduce semantic drift, where clinically precise meaning is distorted during simplification.

We present a controlled study of semantic consistency between semi-synthetic menstrual-health records and the patient-facing summaries LLMs generate from them. Evidence-grounding constraints are expressed as prompt instructions and optionally combined with lay-term definitions retrieved via Retrieval-Augmented Generation (RAG). We evaluate alignment with the source record using automatic evaluation metrics based on Natural Language Inference (NLI) and atomic factuality, alongside readability measures.

Across three LLMs and 4,500 generated summaries, explicit grounding constraints consistently improve alignment with the source record, particularly uncertainty preservation; faithfulness, evaluated against the record and the retrieved definitions, is generally preserved or improved, while effects on omission remain model-dependent. Improved alignment comes at the cost of increased reading difficulty and more technical language. As the records are semi-synthetic and the metrics are not yet validated against expert judgment, these findings characterize semantic consistency rather than clinical validity. They highlight the need to jointly evaluate faithfulness, uncertainty preservation, and readability when developing patient-facing medical LLM systems.

\end{abstract}

\section{Introduction}
The growing integration of wearable and digital health technologies into everyday life has created an expanding stream of longitudinal physiological data (spanning menstrual cycle phases, heart rate variability (HRV), skin temperature, and sleep quality) that is increasingly interpreted by patients themselves rather than by clinicians. Communicating these observations in accessible lay language is therefore a task with direct implications for patient safety and health literacy.

Large Language Models (LLMs) present a compelling mechanism for this translation. Recent work demonstrates that adapted LLMs can match or exceed medical expert performance on standard clinical summarization benchmarks \cite{d5febee1f8294dd79fdfb727baf375bd}, and the field has seen rapid advances in factuality verification, constrained generation, and longitudinal narrative synthesis. Semantic drift, defined as the accumulation of semantic change as medical concepts move from clinical to public discourse, has been shown to be greatest at precisely this transition point \cite{karakkaparambil-james-etal-2026-semantic}. Our work operationalizes this broader phenomenon in the context of automated summarization, introducing a five-category clinical alignment taxonomy and a constrained generation framework to systematically measure and mitigate misalignment across these dimensions.

Existing approaches to health text simplification assume a semantically equivalent reference document. Our setting is fundamentally different, generating patient-facing narratives directly from structured physiological observations, where faithfulness must be verified against raw evidence rather than a reference text. This gap is particularly consequential in menstrual health monitoring, where cyclical signals, intrinsic uncertainty, and the documented consequences of urgency miscalibration make evidence-grounded generation especially critical.
We study this problem in a controlled setting: records are semi-synthetic, generated from a causal model of the menstrual cycle, and alignment is measured with automatic proxy metrics. Our results therefore characterize the semantic consistency of summaries with their source records, not clinical validity on real-world data.

\begin{figure*}[!t]
    \vspace{-0.5em}
    \centering
    \includegraphics[width=0.9\linewidth]{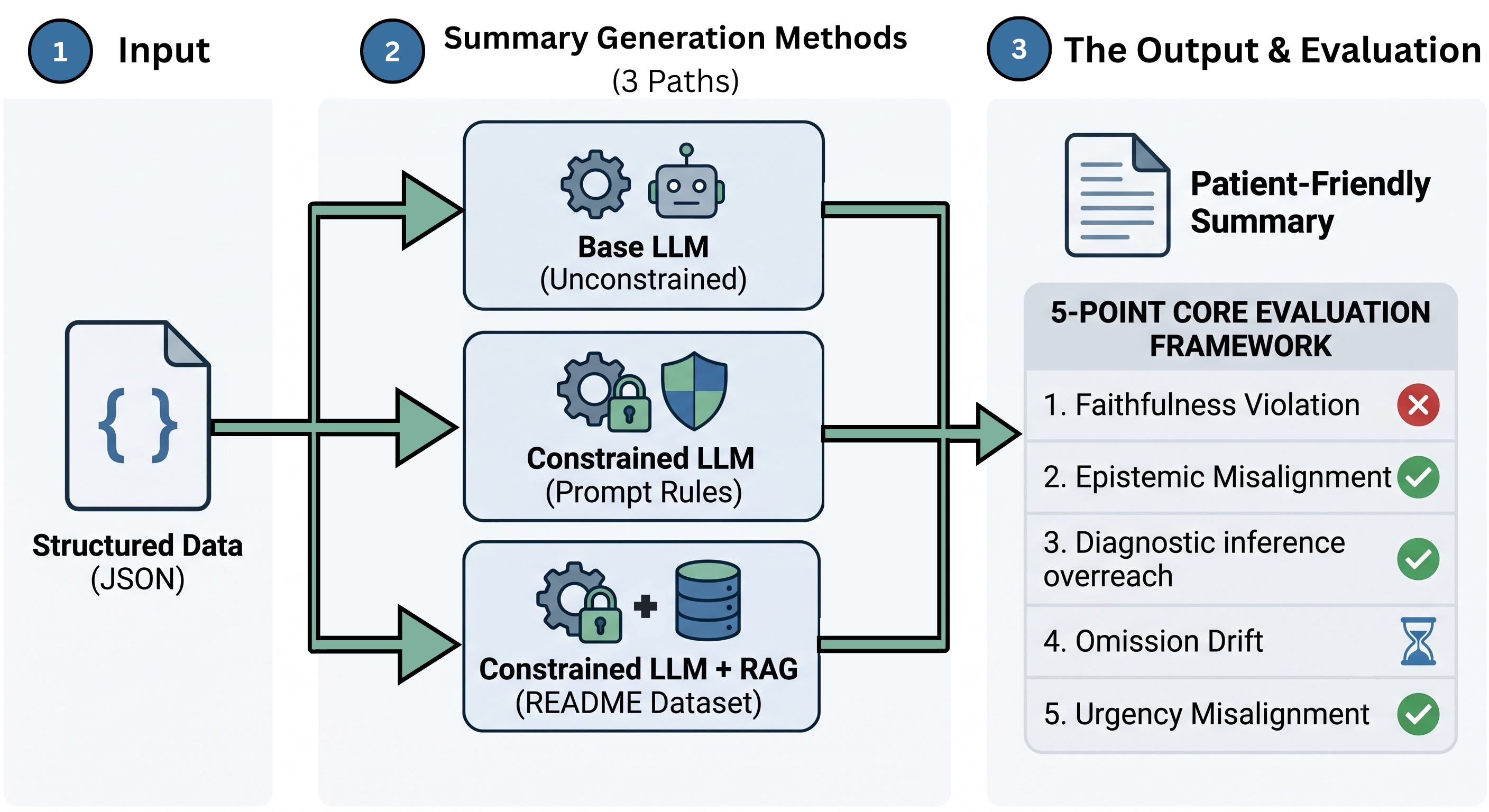}
    \vspace{-0.5em}
    \caption{Overview of the pipeline and five-category semantic alignment evaluation framework.}
    \vspace{-1.0em}
    \label{fig:pipeline}
\end{figure*}

This paper makes the following contributions:
\begin{itemize}
    \item We formalize the structured-to-narrative summarization task for longitudinal physiological records and introduce an evaluation concept using a taxonomy of five semantic drift conditions: Faithfulness Violation, Epistemic Misalignment, Diagnostic Inference Overreach, Omission Drift, and Urgency Misalignment.
    \item We design and implement a framework for the semi-synthetic generation of physiologically motivated menstrual health data.
    \item  We propose an LLM-based generation pipeline combining explicit evidence-grounding constraints, targeting each alignment category in the taxonomy, with RAG that supports simplification through retrieved lay-term definitions. Figure~\ref{fig:pipeline} provides an overview of the proposed framework. 
    \item We systematically evaluate, on semi-synthetic records and with automatic proxy metrics, the alignment of generated patient narratives along the defined taxonomy and derive preliminary insights on favorable configurations of such a setup in menstrual health monitoring.
\end{itemize}
Taken together, this work positions trustworthy simplification as a central and measurable challenge in human-centered digital health.

\section{Related Work}

\subsection{Clinical Summarization and Lay-Language Simplification}

Clinical summarization includes both expert-to-expert tasks, such as radiology impressions and discharge summaries, and expert-to-patient communication, which requires lay-language simplification grounded in health literacy. \cite{d5febee1f8294dd79fdfb727baf375bd} show that adapted LLMs can match or exceed human expert performance across multiple clinical summarization tasks, suggesting that output fluency is less of a bottleneck than semantic fidelity. In longitudinal settings, CLIN-SUMM \cite{souza} demonstrates the value of temporally structured summarization, while \cite{ferrazzi-etal-2026-small} show that supervised fine-tuning and constrained decoding can allow small domain-specific models to outperform much larger ones.
This motivates our use of explicit generation constraints, though we apply them at the prompting level rather than through decoding-time control.
Decoding-time control can guarantee output structure but hardly semantic properties such as hedging; we leave deterministic or tool-based enforcement to future work.

For patient-facing communication, simplification is closely tied to health literacy and self-management outcomes \cite{fitzpatrick2023improving}. Resources such as README \cite{yao-etal-2024-readme} and multi-stage simplification pipelines like S3 support medical-to-lay translation, but prior work consistently highlights a trade-off between readability and fidelity. Simplification can improve accessibility while omitting or distorting clinically important qualifiers, a pattern consistent with semantic drift in the transition from evidence to narrative \cite{karakkaparambil-james-etal-2026-semantic}. Recent work addresses this tension through fidelity-readability optimization \cite{11359644}, and reproductive health further complicates simplification due to indirect or stigmatized language \cite{hua2026openbloomquestionbasedllmtool}. These findings underscore the need for evidence-grounded generation.

However, existing benchmarks focus on summarizing or simplifying \emph{existing} clinical text. They do not address generation directly from structured physiological observations, where faithfulness must be preserved against raw evidence rather than a reference document.

\subsection{NLI-Based Faithfulness Assessment and Hallucination Verification}

Natural Language Inference (NLI) \cite{bowman-etal-2015-large, williams-etal-2018-broad} is widely used for factuality assessment in summarization, especially when combined with atomic fact decomposition \cite{min-etal-2023-factscore, kamoi-etal-2023-wice}. In medical settings, this is particularly relevant because hallucinations often reflect reasoning failures rather than missing knowledge \cite{pal-etal-2023-med}. Benchmarks such as Med-HALT and MedHalu show that instruction tuning can even worsen hallucination behavior, motivating verification-based evaluation.

Clinical NLI was established by MedNLI \cite{romanov-shivade-2018-lessons}, but several limitations remain. Clinical NLI datasets contain annotation artifacts that can inflate performance \cite{herlihy-rudinger-2021-mednli}, NLI models degrade under domain shift \cite{jullien-etal-2024-semeval}, and textual entailment is not equivalent to factual support \cite{mor-lan-levi-2024-exploring}. In addition, atomic-level verification depends heavily on the consistency of fact decomposition in clinical text \cite{munnangi2025factehrdatasetevaluatingfactuality}.

The most relevant paradigm is decompose-then-verify. FActScore \cite{min-etal-2023-factscore} evaluates generated text at the atomic-claim level against trusted evidence, while AlignScore \cite{zha-etal-2023-alignscore} learns a broader alignment function across NLI, QA, and IR tasks. Despite known limitations, these claim-level methods are better suited than surface metrics such as ROUGE for detecting the subtle semantic drift central to our setting.

Overall, prior work has studied constrained generation, factuality verification, and longitudinal summarization largely in isolation. To our knowledge, no work has combined them for structured-to-narrative, patient-facing summarization of wearable health data. Our paper addresses this gap.

\section{Methodology}
\label{sec:methodology}
This work investigates whether evidence-grounded and constrained language generation can improve semantic alignment during the simplification of longitudinal health observations into patient-facing summaries. The framework targets factual consistency, uncertainty calibration, and clinical faithfulness while maintaining accessibility for non-expert users.
\subsection{Task Definition}
Given a structured set of physiological observations derived from longitudinal health monitoring, the task is to generate a patient-oriented natural language summary that remains faithful to the underlying evidence, avoids unsupported medical claims, preserves clinically relevant uncertainty, and is understandable to non-expert users. Formally, let $X = \{x_1, x_2, \ldots, x_n\}$ denote a structured physiological record 
containing observed signals, symptom annotations, and contextual metadata. The objective 
is to generate a lay-language summary $Y$ such that the generated content remains 
semantically aligned with $X$ while minimizing factual distortion. 
The evaluation of this semantic alignment is detailed in Sec.~\ref{sec:evaluation}.

\subsection{Dataset Construction}
\label{sec:dataset}
Due to the limited availability of publicly accessible longitudinal menstrual-health datasets paired with patient-oriented summaries, this study employs a semi-synthetic structured dataset generated via a causal directed acyclic graph (DAG) that encodes known physiological dependencies across the menstrual cycle. The generator produces 500 single-day observations, each drawn from a stochastically determined cycle, following a biologically motivated causal ordering: temporal structure (cycle and luteal-phase length, calibrated to population data \cite{bull2019real,fehring2006variability}) determines phase and cycle day; sleep quality and duration are conditioned on phase, reflecting progesterone-mediated sleep disruption \cite{baker2007sleep}; resting heart rate and HRV trend are jointly conditioned on phase and sleep, capturing both hormonal modulation \cite{https://doi.org/10.1111/psyp.12250} and acute sympathetic override; SpO$_2$ is sampled independently as a homeostatically regulated signal; and symptoms are generated as downstream consequences of these upstream states rather than independent annotations.
Cycle length follows a lognormal distribution matched to the population statistics of \citet{bull2019real} ($\mu{=}3.355$, $\sigma{=}0.155$, clipped to 21--50 days; median ${\approx}28.7$), and luteal length $\mathcal{N}(14.0, 1.5)$ clipped to 10--17 days, as inter-cycle variability lies mainly in the follicular phase \citep{fehring2006variability}. Menses span days 1--5, ovulation is day $L_{\text{cycle}} - L_{\text{luteal}}$, and the cycle day is sampled uniformly. Sleep quality, HRV trend, and temperature trend are phase-conditional categorical variables; sleep duration and resting heart rate are normal given sleep quality and phase, with a 4--8\,bpm RHR penalty and a \textit{decreasing}/\textit{erratic} HRV override under poor sleep; SpO$_2$ is independent; and symptoms follow rules over these variables. All parameters are given in Appendix~\ref{app:datagen}.

To simulate real-world data sparsity, each sample is assigned one of five user profiles (\textit{minimal\_tracker} to \textit{advanced\_wearable}) that determine which fields are observable, probabilistically available, or systematically absent, producing behaviorally mediated structured missingness.
A confidence score, derived from resulting data completeness and phase clarity, provides uncertainty metadata for downstream generation.
The full causal record (pre-masking) is retained as ground truth $X$, enabling evaluation of whether generated summaries faithfully reflect available evidence without hallucinating masked information.

Full details of the generative DAG and an example record are provided in Appendix~\ref{app:datagen}.

\subsection{System Overview}
Our system takes each patient’s structured menstrual-health record (fields such as cycle timing, symptoms, trends, and related clinical context) and generates patient-facing summaries under three generation regimes: \base, \constrained, and \ragconstrained . In the Base setting, the model summarizes directly from the record with minimal control; in the Constrained setting, the same input is generated under explicit clinical communication constraints, implemented as structured instructions within the prompt (for example, factual grounding, uncertainty handling, and nuance-preservation rules, full prompts in Appendix~\ref{app:system_prompts}). Finally, in the RAG-Constrained setting, those constraints are retained while adding retrieved lay-term definitions to support clearer, simplified explanations (see Appendix~\ref{app:rag} for implementation details). Specifically, the role of Retrieval-Augmented Generation (RAG) in this setting is to facilitate text simplification using the README dataset, an extensive collection of over 50,000 unique pairs of medical terms and their lay definitions \cite{yao-etal-2024-readme}.
Together, these constraints enforce evidence-aware summarization over open-ended medical interpretation.
Because the constraints are prompt instructions rather than deterministic decoding rules, compliance is not guaranteed; our metrics measure how far each model actually follows them.

\subsection{Experimental Setup and Baselines}
\label{sec:experimental_setup}
To systematically evaluate the impact of our evidence-grounded generation framework, we benchmarked the task across three distinct tiers of Large Language Models. This selection allows us to observe how model scale affects clinical alignment and constraint adherence:

\begin{itemize}
    \item \textbf{GPT-4o}: Representing a state-of-the-art, large-scale proprietary model \cite{openai2024gpt4technicalreport}.
    \item \textbf{Qwen2.5:14B-Instruct}: Representing a highly capable, mid-sized open-weight model \cite{qwen2025qwen25technicalreport}.
    \item \textbf{Phi-4-mini-4B}: Representing a lightweight, efficient small language model (SLM) suitable for low-latency or on-device health applications \cite{microsoft2025phi4minitechnicalreportcompact}.
\end{itemize}
These models span different scales rather than the current frontier; the framework and metrics are model-agnostic and apply unchanged to newer models.

For our experiments, we sampled 500 records from the semi-synthetic menstrual health dataset. Each record was processed across the three models using three distinct generation configurations, resulting in a total corpus of 4,500 patient-facing summaries. The configurations evaluated include:

\begin{itemize}
    \item \textbf{\base (Vanilla LLM)}: The model directly translates the structured physiological observations into a summary.
    \item \textbf{\constrained}: The model generates a summary utilizing explicit evidence-grounding constraints (e.g., Uncertainty Preservation, Diagnostic Constraint) injected into the prompt, but operates without Retrieval-Augmented Generation (RAG).
    \item \textbf{\ragconstrained }: The full pipeline combining non-parametric retrieval from a curated medical knowledge base (medical term, lay definition) with explicit prompt based clinical constraints.
\end{itemize}

\subsection{Evaluation}
\label{sec:evaluation}
\begin{table*}[t]
\centering
\small
\begin{tabular}{p{2.4cm}p{3.2cm}p{3.6cm}p{1.3cm}p{3.2cm}}
\hline
\textbf{Alignment Condition} & \textbf{Definition} & \textbf{Example} & \textbf{Metric} & \textbf{Related Work} \\ \hline
\addlinespace
Faithfulness Violation & Unsupported physiological claims generated beyond the scope of the structured evidence & Stating elevated HRV when only temperature data is present & Eq.~\ref{eq:faith} & Faithfulness and factuality\cite{maynez-etal-2020-faithfulness} \\
\addlinespace
Epistemic misalignment & Amplification of uncertain observations into definitive statements & ``Your cycle is irregular'' instead of ``your cycle may be irregular'' & Eq.~\ref{eq:epist} & \cite{clark2025epistemic, ayadi-2026-clinical} \\
\addlinespace
Diagnostic inference overreach & Implication of pathology unsupported by the underlying evidence & Suggesting endometriosis from mild cramp annotations & Eq.~\ref{eq:diag} & \cite{zhu2026causalstoriessensortraces} \\
\addlinespace
Omission Drift & Removal of clinically relevant nuance during simplification & Dropping confidence scores or phase uncertainty from the summary & Eq.~\ref{eq:omit} & \cite{oukelmoun-etal-2025-detecting} \\
\addlinespace
Urgency Misalignment & Unnecessarily alarming or over-reassuring language & ``This is completely normal'' for an ambiguous HRV drop & Eq.~\ref{eq:urg} & \\
\addlinespace
\hline
\end{tabular}
\caption{Semantic alignment taxonomy covering five clinical alignment conditions, their definitions, illustrative examples, formal metrics, and connections to related literature concepts.}
\label{tab:drift_taxonomy}
\end{table*}

We define all semantic alignment metrics so that a score of zero indicates perfect alignment and higher values indicate greater drift. Let $X$ denote the structured physiological record and $Y$ the generated patient-facing summary.

\noindent\textbf{Faithfulness violation ($M_{\mathrm{faith}}$).}
Each summary's precomputed \texttt{atomic\_facts} $A$ are checked against a textual premise $P$. For \base{} and \constrained{}, $P = X$,  where $X$ is textualized by concatenating all non-null fields of $X$ (excluding \texttt{sample\_id}) into field-value statements. For \ragconstrained{}, whose prompt explicitly licenses the retrieved lay-term definitions $C_{\mathrm{ret}}$ for explanations, $P = X \cup C_{\mathrm{ret}}$. Since $C_{\mathrm{ret}}$ contains no patient-specific values, patient-specific assertions (e.g., ``your HRV is decreasing'') can only be supported by $X$, whereas general explanations (e.g., ``HRV is the variation in time between heartbeats'') may be supported by $C_{\mathrm{ret}}$. Each claim $a_i \in A$ is tested with \texttt{pritamdeka/PubMedBERT-MNLI-MedNLI} \cite{deka-etal-2023-multiple} via the Hugging Face text-classification pipeline, with $P$ as premise and $a_i$ as hypothesis; with $\mathrm{NLI}(P, a_i) \in \{\text{entailment}, \text{neutral}, \text{contradiction}\}$, a claim is supported only if labeled entailment:
\begin{equation}
\mathbb{I}_{\mathrm{sup}}(a_i) =
\begin{cases}
1 & \text{if } \mathrm{NLI}(P, a_i) = \text{entailment} \\
0 & \text{otherwise}
\end{cases}
\end{equation}

\begin{equation}
M_{\mathrm{faith}} = \frac{1}{|A|}\sum_{i=1}^{|A|}
\left(1 - \mathbb{I}_{\mathrm{sup}}(a_i)\right)
\label{eq:faith}
\end{equation}

\noindent\textbf{Epistemic misalignment ($M_{\mathrm{epist}}$).}
A record bears uncertainty if \texttt{confidence\_score} $<0.60$ or \texttt{missing\_fields} is non-empty. The threshold of $0.60$ was chosen heuristically to operationalize low-confidence observations. Claims from such records are uncertainty-sensitive if they contain cycle- or interpretation-related keywords (e.g., \textit{phase}, \textit{ovulation}, \textit{trend}, \textit{normal}). A claim is definitive if it contains no hedge markers (e.g., \textit{may}, \textit{might}, \textit{likely}) and is asserted directly:
\begin{equation}
M_{\mathrm{epist}}=\frac{1}{|A_{\mathrm{unc}}|}\sum_{a \in A_{\mathrm{unc}}}\mathbb{I}(\mathrm{IsDefinitive}(a)).
\label{eq:epist} 
\end{equation}
Here, $A_{\mathrm{unc}}$ denotes the set of
uncertainty-sensitive claims from records with
a confidence score below 0.60 or at least one
missing field. We set $M_{\mathrm{epist}} = 0$
when $A_{\mathrm{unc}} = \emptyset$.

\noindent\textbf{Diagnostic inference overreach ($M_{\mathrm{diag}}$).}
Unsupported pathology mentions are detected via scispaCy \texttt{en\_ner\_bc5cdr\_md} with medSpaCy negation handling; only non-negated \texttt{DISEASE} mentions are retained as $D_Y$. A summary overreaches if any extracted diagnosis is absent from the textualized source record:
\begin{equation}
M_{\mathrm{diag}}=\mathbb{I}(D_Y \not\subseteq X).
\label{eq:diag} 
\end{equation}
We also report a corrected version excluding \texttt{pregnancy} and \texttt{ectopic pregnancy}, which manual review showed were typically explanatory reproductive-context mentions rather than unsupported diagnoses.

\noindent\textbf{Omission drift ($M_{\mathrm{omit}}$).}
For a fixed salient-field set (cycle timing, phase, ovulation estimate, sleep, resting heart rate, HRV trend, SpO$_2$, temperature trend, symptoms), we generate a reference statement per non-null field and test reversed entailment with the same NLI model (summary as premise, field statement as hypothesis). Unentailed fields count as omitted:
\begin{equation}
M_{\mathrm{omit}}=\frac{\#\text{ omitted salient fields}}{\#\text{ non-null salient fields in }X}.
\label{eq:omit}
\end{equation}
$M_{\mathrm{omit}}$ thus measures strict, surface-level recall of field values via entailment; it can penalize valid paraphrases in which the information is retained (\S\ref{sec:discussion-clinical}).

\noindent\textbf{Urgency misalignment ($M_{\mathrm{urg}}$).}
Unlike the claim-level metrics above, urgency concerns perceived tone rather than entailment. Each summary $Y$ is rated by an LLM judge (Claude-3.5-Haiku \cite{anthropic2024haikuaddendum} , at zero temperature) on a continuous scale from reassuring to alarming, independent of whether that tone is clinically warranted:
\begin{equation}
M_{\mathrm{urg}} = \mathrm{Judge}(Y) \in [0,1].
\label{eq:urg}
\end{equation}
$M_{\mathrm{urg}}$ therefore measures perceived tone (0 = reassuring, 1 = alarming), not whether that tone is clinically appropriate for the record; unlike the other metrics, it has no fixed target value.

\noindent\textbf{Readability, lexical diversity, and jargon.}
We assess the fidelity-accessibility trade-off along three axes. \\
\textit{Grade-level readability}: Simple Measure of Gobbledygook (SMOG),
Flesch-Kincaid Grade Level (FKGL), and
Automated Readability Index (ARI), plus Dale-Chall (vocabulary familiarity, largely length-independent); since the three grade-level indices are highly correlated, FKGL is our primary endpoint, with SMOG and ARI as robustness checks.\\
\textit{Lexical and syntactic complexity}: Measure of Textual Lexical Diversity (MTLD)
 (forward-backward averaged, factor threshold 0.72) for lexical variety, and a fixed spaCy parser for mean dependency-tree depth and subordinate-clause ratio, applied identically across conditions; word counts are tracked to detect verbosity inflation.\\
\textit{Medical jargon}: scispaCy with a UMLS linker identifies terms resolving to a documented UMLS concept, reported as density (entities per 100 words), raw entity count, and entities per sentence.\subsection{Statistical Procedure}
\label{sec:readability-stats}
 
For each metric we run paired Wilcoxon signed-rank tests \cite{wilcoxon1992individual}  , since the
design is fully matched and several metrics are not normally
distributed. We report the paired mean and median difference, a paired
Cohen's $d$ (mean difference over the standard deviation of the paired
differences), and the matched-pairs rank-biserial correlation as a
non-parametric effect size. Two families of contrasts are tested
separately: Within-model comparisons across the three summary types,
and within-summary-type comparisons across the three models. Because
n~$=500$ pairs per test, most contrasts reach statistical significance;
we therefore treat effect size, not the raw $p$-value, as the basis for
interpretation. $p$-values are corrected for multiple comparisons with
the Benjamini-Hochberg procedure, applied separately within each of the
two contrast families.

\section{Results}
\label{sec:results}

Table~\ref{tab:clinical_alignment_combined} summarizes the five alignment metrics together with the main readability and jargon measures for all 4,500 summaries, per model and pooled across models. FKGL is the primary grade-level endpoint (\S\ref{sec:evaluation}); Dale-Chall, MTLD, dependency depth, and subordinate-clause ratio are interpreted as distinct constructs.

\subsection{Clinical alignment}

\textbf{Faithfulness violation.} Table~\ref{tab:clinical_alignment_combined} reports mean faithfulness violation rates across
models and settings (premise $P$ as defined in Eq.~\ref{eq:faith}). For all three
models, $M_{\text{faith}}$ is lowest or comparable in the RAG-constrained condition relative
to constrained: GPT-4o improves further under retrieval, Qwen2.5-14B is
essentially unchanged, and Phi-4-mini shows a small increase. All three models remain well below their base levels.
Pooled across models, mean $M_{\mathrm{faith}}$
decreases monotonically from 0.242 to 0.178,
representing a relative reduction of 26.4\%;
 averaged over settings, Qwen2.5-14B performs best overall (0.172), followed by
GPT-4o (0.201) and Phi-4-mini (0.235).

\paragraph{Epistemic misalignment.} As shown in Table~\ref{tab:clinical_alignment_combined}, $M_{\mathrm{epist}}$ decreases monotonically from \texttt{base} to \texttt{constrained} to \texttt{rag\_constrained} for all three models and pooled across models (Table~\ref{tab:clinical_alignment_combined}). GPT-4o
 achieves the lowest final value; Phi-4-mini and Qwen2.5-14B show the same monotonic pattern. Median $M_{\mathrm{epist}}$ is 0.0 in every model-by-setting cell, so the mean is driven by a smaller subset of problematic cases. Notably, this improvement occurs despite the average number of uncertainty-sensitive claims rising across settings (pooled: 1.17 $\rightarrow$ 1.76 $\rightarrow$ 2.70), indicating that explicit uncertainty-preservation instructions and retrieved explanatory support help models state uncertain observations more cautiously rather than more definitively.

\paragraph{Diagnostic inference overreach.} The raw diagnostic-overreach metric suggested substantial monotonic increases under constraints and retrieval; manual inspection traced nearly all flagged cases to \emph{pregnancy} or \emph{ectopic pregnancy} mentions used in fertility-window or ovulation explanations rather than as unsupported diagnoses (discussed further in \S\ref{sec:discussion-clinical}). After excluding these two terms, overreach collapses to near zero in every cell (at most 0.006; Table~\ref{tab:clinical_alignment_combined}).

\paragraph{Omission drift.} Omission drift shows more heterogeneous behavior. GPT-4o improves monotonically, with a median of 0.0 under \texttt{rag\_constrained}. Qwen2.5-14B performs best under \texttt{constrained} generation and rises slightly under retrieval, remaining below its \texttt{base} value. Phi-4-mini is the only model that worsens monotonically.
 Pooled across models, omission drift declines only modestly (Table~\ref{tab:clinical_alignment_combined}), masking these model-specific differences;
 we relate Phi-4-mini's reversal to its more discursive output style in \S\ref{sec:discussion-clinical}.

\paragraph{Urgency misalignment.} Mean $M_{\mathrm{urg}}$ increases significantly from \texttt{base} to \texttt{constrained} for all three models (e.g., GPT-4o: 0.170 $\rightarrow$ 0.227, $d=-0.61$; all $p_{\mathrm{FDR}}<10^{-5}$), then drops significantly from \texttt{constrained} to \texttt{rag\_constrained} for every model ($d=0.37$–$0.45$, all $p_{\mathrm{FDR}}<10^{-13}$), returning GPT-4o and Phi-4-mini to \texttt{base} levels while Qwen2.5-14B only partially recovers. Across settings, GPT-4o is consistently the least alarming model, though this gap narrows once retrieval is added.

\begin{table*}[t]
\centering
\small
\begin{tabular}{ll ccccc ccc}
\toprule
 & & \multicolumn{5}{c}{\textbf{Alignment}} & \multicolumn{3}{c}{\textbf{Readability / jargon}} \\
\cmidrule(lr){3-7}\cmidrule(lr){8-10}
\textbf{Model} & \textbf{Setting} & \textbf{$M_{\text{faith}}\downarrow$} & \textbf{$M_{\text{epist}}\downarrow$} & \textbf{$M_{\text{diag}}\downarrow$} & \textbf{$M_{\text{omit}}\downarrow$} & \textbf{$M_{\text{urg}}$} & \textbf{FKGL} & \textbf{D-C} & \textbf{Med.\,Ent.} \\
\midrule
GPT-4o
    & Base            & 0.264 & 0.293 & 0.000 & 0.219 & 0.170 & 7.38 & 9.67 & 16.63 \\
    & Constrained     & 0.186 & 0.243 & 0.000 & 0.195 & 0.227 & 9.93 & 10.60 & 26.07 \\
    & RAG-Constrained & 0.152 & 0.237 & 0.000 & 0.130 & 0.179 & 10.31 & 10.54 & 32.29 \\
\midrule
Qwen2.5-14B
    & Base            & 0.190 & 0.299 & 0.000 & 0.176 & 0.247 & 7.64 & 10.07 & 18.38 \\
    & Constrained     & 0.165 & 0.275 & 0.000 & 0.147 & 0.270 & 9.72 & 11.01 & 26.26 \\
    & RAG-Constrained & 0.162 & 0.251 & 0.000 & 0.167 & 0.199 & 10.39 & 10.92 & 32.44 \\
\midrule
Phi-4-mini
    & Base            & 0.273 & 0.298 & 0.004 & 0.264 & 0.237 & 10.63 & 9.84 & 24.28 \\
    & Constrained     & 0.212 & 0.270 & 0.000 & 0.273 & 0.263 & 12.18 & 10.44 & 33.61 \\
    & RAG-Constrained & 0.220 & 0.255 & 0.006 & 0.286 & 0.181 & 12.32 & 10.38 & 40.96 \\
\midrule
Pooled
    & Base            & 0.242 & 0.297 & 0.001 & 0.220 & 0.218 & 8.55 & 9.86 & 19.76 \\
    & Constrained     & 0.188 & 0.263 & 0.000 & 0.205 & 0.253 & 10.61 & 10.68 & 28.65 \\
    & RAG-Constrained & 0.178 & 0.248 & 0.002 & 0.194 & 0.186 & 11.01 & 10.61 & 35.23 \\
\bottomrule
\end{tabular}
\caption{Alignment, readability, and jargon metrics by model and generation setting, ordered by model scale (large to small); \emph{Pooled} averages the three models. $M_{\text{faith}}$: faithfulness violation rate (premise $P$, Eq.~\ref{eq:faith}); $M_{\text{epist}}$: epistemic misalignment; $M_{\text{diag}}$: diagnostic inference overreach (corrected, excluding pregnancy/ectopic pregnancy mentions); $M_{\text{omit}}$: omission drift; $M_{\text{urg}}$: urgency (tone from 0 = reassuring to 1 = alarming; no fixed target). Alignment metrics range $[0,1]$; except for $M_{\text{urg}}$, lower values indicate better alignment. FKGL: Flesch-Kincaid grade level; D-C: Dale-Chall score; Med.\,Ent.: medical entities per summary Higher FKGL and D-C scores indicate greater reading
difficulty, while higher Med. Ent. values indicate
more medical terminology. Full readability and jargon results in Appendices~\ref{sec:readability} and~\ref{sec:jargon}.}
\label{tab:clinical_alignment_combined}
\end{table*}

\subsection{Readability and stylistic trade-offs}

\paragraph{Grade-level readability and syntactic complexity.} Across all three models, readability worsens under added control: all three grade-level metrics increase monotonically or near-monotonically from \texttt{base} to \texttt{constrained} to \texttt{rag\_constrained} (Table~\ref{tab:clinical_alignment_combined}; ARI in Table~\ref{tab:readability}), with FKGL as the primary endpoint.
 For GPT-4o, the largest shift occurs between \texttt{base} and \texttt{constrained} (FKGL paired $d = 1.39$; ARI $d = 1.22$; Dale-Chall $d = 1.23$; all $p_{\mathrm{FDR}} < 10^{-70}$), while the further increase from \texttt{constrained} to \texttt{rag\_constrained} is modest for FKGL ($d = 0.26$) and negligible for ARI ($d = -0.08$, n.s.\ after correction) and Dale-Chall ($d = -0.09$, significant but trivial). Dependency-tree depth follows the same pattern: a large increase from \texttt{base} to \texttt{constrained} ($d = 0.90$) with essentially no further change under retrieval ($d = 0.01$, n.s.). This suggests explicit grounding constraints, rather than retrieval itself, are the main driver of increased reading difficulty. Table~\ref{tab:readability} shows the same qualitative pattern across models: Phi-4-mini produces the highest grade-level outputs under all settings, GPT-4o occupies the middle range, and Qwen2.5-14B is typically lowest on FKGL and ARI once constraints are applied.

\paragraph{Lexical diversity.}
As shown in Table~\ref{tab:readability},
using MTLD (to assess lexical variety independent of length), GPT-4o shows a clear decline from \texttt{base} to \texttt{rag\_constrained}, modest between \texttt{base} and \texttt{constrained} ($d = -0.26$) and substantially larger between \texttt{base} and \texttt{rag\_constrained} ($d = -0.63$). Phi-4-mini declines similarly, from a much higher starting point. Qwen2.5-14B is comparatively stable, even rising slightly from \texttt{base} to \texttt{constrained} before returning near baseline under retrieval. The lexical-diversity cost of constrained generation is therefore model-dependent rather than universal.

\paragraph{Medical jargon.} Medical-jargon usage shows a more nuanced trade-off than a single measure captures. As shown in Tables~\ref{tab:clinical_alignment_combined} and~\ref{tab:jargon_metrics}, raw medical entity count rises consistently from \texttt{base} to \texttt{rag\_constrained} for every model, with large effects for GPT-4o (\texttt{constrained} minus \texttt{base}: $d=1.89$; \texttt{rag\_constrained} minus \texttt{base}: $d=2.24$). By contrast, jargon density (entities per 100 words) declines across settings for all models, since total word count grows.
Table~\ref{tab:qual_examples} illustrates the shift in length, terminology, and hedging for one record; examples for all models and settings are given in Appendix~\ref{app:qual_examples}.

\begin{table*}[t]
\centering
\small
\begin{tabular}{p{0.30\textwidth}p{0.30\textwidth}p{0.30\textwidth}}
\toprule
\multicolumn{3}{p{0.95\textwidth}}{\textbf{Input record (MH-0094, as shown to the model):} Day 8, follicular; cycle length 30 days; sleep: good, 7.97\,h; RHR 65.19\,bpm; HRV trend: stable; SpO$_2$ 98.5\%; symptoms: none; note: ``Feeling a bit off today [\ldots]''; confidence 0.86.} \\
\midrule
\textbf{Base} & \textbf{Constrained} & \textbf{RAG-Constrained} \\
\midrule
You're on day 8 of your cycle, currently in the follicular phase, with a typical cycle length of about 30 days. Your sleep quality was good, and you got nearly 8 hours of rest, which is great! [\ldots] Consider taking some downtime and reducing screen time to recharge. Keep up the good work caring for yourself!
&
\textbf{Your Current Status} Today is day 8 of your cycle, and you're in the follicular phase. [\ldots] \textbf{What Your Data is Showing} [\ldots] Your resting heart rate is at 65 beats per minute, and your heart rate variability trend \textit{appears} stable, both of which \textit{suggest} your body is maintaining balance. [\ldots]
&
\textbf{Your Current Status} You are currently on day 8 of your menstrual cycle, which places you in the follicular phase. This is the phase that occurs after your period and before ovulation. [\ldots] \textbf{Symptom Context} [\ldots] The data provided has a confidence score of 0.86, which \textit{suggests} a fairly reliable snapshot of your current state. [\ldots] \\
\midrule
\multicolumn{3}{l}{\textit{Full summary:} 92 / 124 / 191 words; FKGL 8.1 / 9.0 / 10.0; 17 / 26 / 37 medical entities} \\
\bottomrule
\end{tabular}
\caption{Excerpts of GPT-4o summaries of one record under the three generation settings ([\ldots] marks omitted sentences). Section headers are the model's own; hedge markers used by $M_{\mathrm{epist}}$ are in italics. Full summaries of all three models are given in Appendix~\ref{app:qual_examples}.}
\label{tab:qual_examples}
\end{table*}

\section{Discussion}

\subsection{Readability}
\label{sec:readability-discussion}
 
Taken together, these results indicate that the evidence-grounding
mechanisms evaluated for their effect on factual alignment in
Section~\ref{sec:methodology} carry a measurable cost in accessibility.
Grade-level readability increases substantially once explicit
constraints are applied, largely independent of whether retrieval is
also used, while lexical diversity decreases and domain-term mentions
increase in absolute terms even as their density per word declines.
This is consistent with the readability-fidelity tension noted in
constrained clinical summarization more broadly \cite{11359644},
and suggests that the drift-reduction constraints introduced in this
work should be evaluated jointly against both factual alignment and
reading difficulty rather than either in isolation. We treat this joint
analysis, correlating the per-sample drift metrics of
Table~\ref{tab:drift_taxonomy} against the per-sample readability
measures reported here, as a direct extension of the present results.
One caveat applies to how we measure readability: FKGL, ARI, and SMOG are based on sentence length and syllable counts, not on how conceptually difficult a word actually is. This means a technical term like "ptosis" can score as easier than an everyday word like "congratulations" simply because it has fewer syllables. This matters for our RAG results as FKGL increases modestly from Constrained to RAG-Constrained (d=0.26) despite stable sentence length, driven mainly by near-doubled medical-term counts. This suggests the extra difficulty these formulas detect comes mainly from added vocabulary, not from harder sentence structure, especially since RAG also supplies a lay definition alongside each term. This does not remove the broader readability cost we report above, but it suggests the part of that cost caused specifically by RAG may be smaller than the raw grade-level scores suggest.

\subsection{Clinical Alignment}
\label{sec:discussion-clinical}

Table~\ref{tab:clinical_alignment_combined} reports all four clinical alignment metrics across models and generation settings.
Across the four metrics, only epistemic misalignment improves consistently across all three models as generation moves from \textit{base}  to \textit{Constrained}  to \textit{RAG-Constrained}. The other three metrics deviate for distinct, explainable reasons.

\textbf{Retrieved context must be part of the faithfulness premise.}
Evaluated against $X$ alone, valid explanations drawn from retrieved definitions (e.g., a plain-language definition of ``HRV'') would be flagged as unsupported. With $P = X \cup C_{\mathrm{ret}}$
(Eq.~\ref{eq:faith}), mean $M_{faith}$ for \textit{rag\_constrained} is significantly lower
than with $P = X$ for all three models ($d = 0.76$--$0.80$, all $p < 10^{-40}$), and is
significantly \emph{lower} than \textit{constrained} for GPT-4o ($d = 0.284$, $p = 1.07 \times 10^{-10}$)
and statistically indistinguishable for Phi-4-mini and Qwen2.5-14B ($p > 0.15$).

Diagnostic overreach is almost entirely a vocabulary artifact. Its apparent rise under RAG is fully explained by terms like "pregnancy" and "ectopic pregnancy." Pregnancy is a physiological state and not a true pathology that legitimately surfaces when explaining fertility, which RAG actively supplies. Removing these terms drops overreach to $\le$0.6 percent for every model and condition, suggesting that the models almost never suggest genuinely unsupported pathologies. We therefore report the corrected figures as the
primary result, distinguishing reproductive-context
mentions from unsupported diagnostic assertions.

Omission drift is model-dependent: GPT-4o benefits from the full pipeline, Qwen2.5-14B from constraints alone, and Phi-4-mini uniquely worsens with more grounding. Phi-4-mini's regression aligns with its readability profile; it adds discursive, explanatory prose rather than restating specific field values outright (e.g., "Resting HR: 71.3"). Because the omission metric checks for direct statements of salient fields, Phi-4-mini fails the strict entailment check even when the underlying information is not truly lost. This indicates the metric is sensitive to surface realization and stylistic differences.
Model rankings on coverage should therefore be interpreted with caution.

These results complicate a simple "more grounding is better" narrative. Explicit constraints alone account for most metric improvements. Retrieval's marginal contribution is genuinely positive for epistemic calibration and, with retrieved context included in the premise, largely positive or neutral for faithfulness but
 its effect on omission remains mixed, and its apparent cost for diagnostic overreach is a vocabulary artifact rather than a real effect.

\section{Conclusion}

On semi-synthetic records and as measured by automatic proxy metrics, we find that evidence-grounded constrained generation improves alignment with the source record along
specific dimensions.
The clearest and most consistent effect across all three models is
reduced epistemic misalignment. Retrieval further improves faithfulness for GPT-4o and is
faithfulness-neutral for the other two models, while its effects on omission and urgency
remain smaller and model-dependent. At the same time, constrained and RAG-constrained
outputs become harder to read, often less lexically diverse, and heavier in medical
terminology.

These results suggest that safer health summarization is not a single-metric problem. Systems must be evaluated jointly for faithfulness, uncertainty preservation, coverage, and readability. 
This work provides a controlled testbed for studying this challenge; confirming these findings on real-world records and against expert judgment remains necessary.
More broadly, it contributes a structured-to-narrative evaluation setting and a semantic alignment framework for studying how LLMs turn physiological data into patient-readable explanations.

\section*{Limitations}

Our study has several limitations that should inform how these results are interpreted.

\paragraph{No human validation of automatic metrics.} All five clinical alignment metrics rely on automatic tools: an NLI model (PubMedBERT-MNLI-MedNLI) for faithfulness violation and omission drift, a named entity recognizer with negation handling for diagnostic inference overreach, and a single LLM judge for urgency misalignment. While these are established approaches in the factuality literature, prior work has documented that clinical NLI models contain annotation artifacts and degrade under domain shift, and that textual entailment does not always correspond to factual support. We do not have a human-annotated subsample to validate any of the five metrics against clinician or expert judgment, and we treat this as the most important direction for follow-up work.
Until such validation is available, the five metrics should be read as proxies for alignment with the source record, not as measures of clinical safety. In particular, omission drift measures strict surface-level recall and can penalize valid paraphrases, and urgency misalignment measures tone rather than its clinical appropriateness.
\paragraph{Single LLM judge for urgency.} Urgency misalignment is scored by a single model (Claude 3.5 Haiku) at a fixed temperature, without inter-rater reliability checks, comparison to human raters, or an ensemble of judges. Given that this metric concerns tone rather than propositional content, it is more subjective than the other four and may be more sensitive to idiosyncrasies of the judge model.

\paragraph{Synthetic data without clinician review.} Our dataset is generated from a causal DAG that encodes documented physiological relationships between cycle phase, sleep, and autonomic measures, and includes internal coherence checks (Appendix \ref{app:datagen}). The DAG structure and its parameters are our own modeling choices, and neither the generated records nor the associated ground truth have been reviewed by a clinician for biological realism. We note that this limitation is partially mitigated by the nature of our evaluation: our alignment metrics measure whether a generated summary $Y$ is semantically faithful to its own source record $X$, not whether $X$ itself is a biologically accurate depiction of the menstrual cycle. In this sense the task is closer to a semantic consistency analysis than a biological validity analysis, and the alignment scores we report do not depend on the DAG being an accurate model of physiology. Clinician review of the DAG would therefore primarily affect the realism of the input distribution rather than the validity of the alignment measurements themselves. That said, this gap still matters for generalization as real-world self-tracked data is likely to be noisier, more idiosyncratic, and structured differently than data produced by a hand-specified generative model, and our results should be understood as characterizing model behavior on this particular semi-synthetic input distribution rather than on real user data.
In particular, the generator does not model clinically important sources of heterogeneity such as irregular cycles or hormonal contraceptive use. Each record is also a single-day observation whose longitudinal context (e.g., cycle length and HRV or temperature trends) is itself produced by the DAG rather than observed.

\paragraph{Narrow instantiation of retrieval.} In our RAG-Constrained setting, retrieval is limited to lay-term definitions drawn from the README dataset and is not used to supply additional clinical evidence beyond the structured record. Our findings about the marginal contribution of retrieval should therefore be understood as specific to this specific use of RAG, and may not generalize to settings where retrieval is used to bring in additional clinical context.
Even this narrow use would require curating and maintaining the lay-term knowledge base (Appendix~\ref{app:rag_kb}).

\paragraph{Model selection and prompt-only constraints.} The three evaluated models span different scales but do not include the most recent frontier models; while the framework is model-agnostic, the reported effects may differ for newer models. Moreover, our constraints are expressed only as prompt instructions and are not enforced deterministically (e.g., through constrained decoding or tool-based verification), so compliance is not guaranteed.

\paragraph{No external baseline system.} As discussed in the main text, we are not aware of prior systems that perform structured-to-narrative generation in this domain, so we rely on internal ablations (Base, Constrained, RAG-Constrained) rather than comparison to an external baseline. This limits our ability to situate the absolute performance of our pipeline relative to other possible approaches.

\paragraph{Domain and demographic scope.} Our experiments focus exclusively on menstrual health monitoring using a semi-synthetic dataset. We do not evaluate other physiological domains, and the generative model does not represent the full diversity of menstrual cycle presentations, medical conditions, or demographic variation present in real populations. Findings may not transfer directly to other health domains or to more clinically atypical records.

\section*{Ethical Considerations}

\paragraph{Data and privacy.} All experiments in this paper use a semi-synthetic dataset generated from a causal statistical model (Appendix \ref{app:datagen}). No real patient records, wearable device data, or personally identifiable health information were collected, accessed, or used at any stage of this work. This eliminates direct privacy risk from our experiments, though it also means our findings should not be taken as validated on real-world data distributions.

\paragraph{Risk of misuse or overreliance.} This work studies failure modes, including faithfulness violation, epistemic misalignment, diagnostic inference overreach, omission drift, and urgency misalignment, in LLM-generated health summaries. Our intent is diagnostic and evaluative: to make these failure modes measurable so that they can be mitigated, not to present any of the three tested generation configurations as ready for deployment in a real patient-facing product. None of the metrics or pipelines described here have been clinically validated, and none should be used, as-is, to make or support real health or diagnostic decisions. In particular, our diagnostic inference overreach metric is intended to \emph{detect} unsupported pathology mentions for research purposes; it is not a clinical safety filter and should not be relied on as one.

\paragraph{Reproductive health context.} Menstrual health monitoring involves personal and sometimes stigmatized health information. We chose this domain because cyclical signals, intrinsic physiological uncertainty, and the documented consequences of urgency miscalibration make it a particularly informative and consequential setting for studying evidence-grounded generation, not because we treat it as a low-stakes proxy for health tasks more broadly. We encourage readers to interpret our results as evidence for the general difficulty of structured-to-narrative health summarization, rather than as specific claims about menstrual health applications being closer to solved than other clinical domains.

\paragraph{Automatic and LLM-based evaluation.} Several of our metrics rely on automatic NLI models, named entity recognition, and an LLM-as-judge (for urgency misalignment). These are research instruments for comparing generation configurations at scale and have known limitations, including domain shift sensitivity and lack of validated correspondence to clinician judgment (see Limitations). They should not be interpreted as clinically validated safety metrics, and results derived from them should not be used to certify any system as safe for patient-facing use.

\paragraph{Downstream use.} We release our dataset generation code (Appendix \ref{app:datagen}) to support reproducibility and further research into evidence-grounded health summarization. We do not release, and did not train, any model intended for deployment. We encourage future work building on this framework to include clinician review and human evaluation of generated summaries before considering any real-world patient-facing application.

\paragraph{Use of AI assistants.} AI assistants were used for coding, code review, and improving the writing.

\section*{Acknowledgments}
We thank Reinhard Stolle and Annette Hautli-Janisz for their valuable guidance, insightful discussions, and continuous support throughout this project. 

This work was supported by the Free State of Bavaria within the DSgenAI project under Grant No.~RMF-SG20-3410-2-18-4.

\bibliography{custom.bib}

@inproceedings{min-etal-2023-factscore,
    title = "{FA}ct{S}core: Fine-grained Atomic Evaluation of Factual Precision in Long Form Text Generation",
    author = "Min, Sewon  and
      Krishna, Kalpesh  and
      Lyu, Xinxi  and
      Lewis, Mike  and
      Yih, Wen-tau  and
      Koh, Pang  and
      Iyyer, Mohit  and
      Zettlemoyer, Luke  and
      Hajishirzi, Hannaneh",
    editor = "Bouamor, Houda  and
      Pino, Juan  and
      Bali, Kalika",
    booktitle = "Proceedings of the 2023 Conference on Empirical Methods in Natural Language Processing",
    month = dec,
    year = "2023",
    address = "Singapore",
    publisher = "Association for Computational Linguistics",
    url = "https://aclanthology.org/2023.emnlp-main.741/",
    doi = "10.18653/v1/2023.emnlp-main.741",
    pages = "12076--12100"
}

@article{d5febee1f8294dd79fdfb727baf375bd,
title = "Adapted large language models can outperform medical experts in clinical text summarization",
author = "Van Veen, Dave and \{Van Uden\}, Cara and Louis Blankemeier and Jean-Benoit Delbrouck and Asad Aali and Christian Bluethgen and Anuj Pareek and Malgorzata Polacin and Reis, \{Eduardo Pontes\} and Anna Seehofnerov{\'a} and Nidhi Rohatgi and Poonam Hosamani and William Collins and Neera Ahuja and Langlotz, \{Curtis P\} and Jason Hom and Sergios Gatidis and John Pauly and Chaudhari, \{Akshay S\}",
note = "{\textcopyright} 2024. The Author(s), under exclusive licence to Springer Nature America, Inc.",
year = "2024",
month = apr,
doi = "10.1038/s41591-024-02855-5",
language = "English",
volume = "30",
pages = "1134--1142",
journal = "Nature Medicine",
issn = "1078-8956",
publisher = "Nature Research",
number = "4",
}

@inproceedings{karakkaparambil-james-etal-2026-semantic,
    title = "Semantic Echo Pathways ({SEP}): Tracing How Medical Language Propagates and Transforms",
    author = "Karakkaparambil James, Charu  and
      Monteiro, Marcio  and
      Fellenz, Sophie",
    editor = {Danilova, Vera  and
      Kurfal{\i}, Murathan  and
      S{\"o}derfeldt, Ylva  and
      Reed, Julia  and
      Burchell, Andrew},
    booktitle = "Proceedings of the 1st Workshop on Linguistic Analysis for Health ({H}ea{L}ing 2026)",
    month = mar,
    year = "2026",
    address = "Rabat, Morocco",
    publisher = "Association for Computational Linguistics",
    url = "https://aclanthology.org/2026.healing-1.5/",
    doi = "10.18653/v1/2026.healing-1.5",
    pages = "55--66",
    ISBN = "979-8-89176-367-8"
}

@ARTICLE{11359644,
  author={Verma, Deepika and Darlan, Daison and Mallipeddi, Rammohan},
  journal={IEEE Access}, 
  title={Multi-Objective Optimization of Large Language Model Summaries for Clinical Text Narratives}, 
  year={2026},
  volume={14},
  number={},
  pages={13597-13609},
  doi={10.1109/ACCESS.2026.3656535}}

@inproceedings{yao-etal-2024-readme,
    title = "{README}: Bridging Medical Jargon and Lay Understanding for Patient Education through Data-Centric {NLP}",
    author = "Yao, Zonghai  and
      Kantu, Nandyala Siddharth  and
      Wei, Guanghao  and
      Tran, Hieu  and
      Duan, Zhangqi  and
      Kwon, Sunjae  and
      Yang, Zhichao  and
      Yu, Hong",
    editor = "Al-Onaizan, Yaser  and
      Bansal, Mohit  and
      Chen, Yun-Nung",
    booktitle = "Findings of the Association for Computational Linguistics: EMNLP 2024",
    month = nov,
    year = "2024",
    address = "Miami, Florida, USA",
    publisher = "Association for Computational Linguistics",
    url = "https://aclanthology.org/2024.findings-emnlp.737/",
    doi = "10.18653/v1/2024.findings-emnlp.737",
    pages = "12609--12629"
}

@misc{hua2026openbloomquestionbasedllmtool,
      title={"OpenBloom": A Question-Based LLM Tool to Support Stigma Reduction in Reproductive Well-Being}, 
      author={Ashley Hua and Adya Daruka and Yang Hong and Sharifa Sultana},
      year={2026},
      eprint={2602.00243},
      archivePrefix={arXiv},
      primaryClass={cs.HC},
      url={https://arxiv.org/abs/2602.00243}, 
}

@inproceedings{pal-etal-2023-med,
    title = "{M}ed-{HALT}: Medical Domain Hallucination Test for Large Language Models",
    author = "Pal, Ankit  and
      Umapathi, Logesh Kumar  and
      Sankarasubbu, Malaikannan",
    editor = "Jiang, Jing  and
      Reitter, David  and
      Deng, Shumin",
    booktitle = "Proceedings of the 27th Conference on Computational Natural Language Learning (CoNLL)",
    month = dec,
    year = "2023",
    address = "Singapore",
    publisher = "Association for Computational Linguistics",
    url = "https://aclanthology.org/2023.conll-1.21/",
    doi = "10.18653/v1/2023.conll-1.21",
    pages = "314--334"
}

@inproceedings{ferrazzi-etal-2026-small,
    title = "Small {LLM}s for Medical {NLP}: A Systematic Analysis of Few-Shot, Constraint Decoding, Fine-Tuning and Continual Pre-Training in {I}talian",
    author = "Ferrazzi, Pietro  and
      Franzin, Mattia  and
      Lavelli, Alberto  and
      Magnini, Bernardo",
    editor = "Piperidis, Stelios  and
      Bel, N{\'u}ria  and
      van den Heuvel, Henk  and
      Ide, Nancy  and
      Krek, Simon  and
      Toral, Antonio",
    booktitle = "Proceedings of the Fifteenth Language Resources and Evaluation Conference",
    month = may,
    year = "2026",
    address = "Palma de Mallorca, Spain",
    publisher = "ELRA Language Resource Association",
    url = "https://aclanthology.org/2026.lrec-1.192/",
    doi = "10.63317/2c6cyoc8xohu",
    pages = "2440--2457"
}

@article {souza,
	author = {D{\textquoteright}souza, Valentina and Azhir, Alaleh and Pace, Danielle F. and Friedman, Samuel and Nargesi, Arash and Naumann, Tristan and Anderson, Christopher D. and Atlas, Steven J. and Hung, Judy and Maddah, Mahnaz},
	title = {CLIN-SUMM: Temporal Summarization of Longitudinal Clinical Notes},
	elocation-id = {2025.11.28.25341233},
	year = {2025},
	doi = {10.64898/2025.11.28.25341233},
	publisher = {Cold Spring Harbor Laboratory Press},
	URL = {https://www.medrxiv.org/content/early/2025/12/02/2025.11.28.25341233},
	eprint = {https://www.medrxiv.org/content/early/2025/12/02/2025.11.28.25341233.full.pdf},
	journal = {medRxiv}
}

@inproceedings{zha-etal-2023-alignscore,
    title = "{A}lign{S}core: Evaluating Factual Consistency with A Unified Alignment Function",
    author = "Zha, Yuheng  and
      Yang, Yichi  and
      Li, Ruichen  and
      Hu, Zhiting",
    editor = "Rogers, Anna  and
      Boyd-Graber, Jordan  and
      Okazaki, Naoaki",
    booktitle = "Proceedings of the 61st Annual Meeting of the Association for Computational Linguistics (Volume 1: Long Papers)",
    month = jul,
    year = "2023",
    address = "Toronto, Canada",
    publisher = "Association for Computational Linguistics",
    url = "https://aclanthology.org/2023.acl-long.634/",
    doi = "10.18653/v1/2023.acl-long.634",
    pages = "11328--11348"
}

@inproceedings{ayadi-2026-clinical,
    title = "The Clinical Fingerprint: Comparing the Rhetorical Integrity and Epistemic Safety of Human Physicians and Large Language Models",
    author = "Ayadi, Bayram",
    editor = "Baez Santamaria, Selene  and
      Somayajula, Sai Ashish  and
      Yamaguchi, Atsuki",
    booktitle = "Proceedings of the 19th Conference of the {E}uropean Chapter of the {A}ssociation for {C}omputational {L}inguistics (Volume 4: Student Research Workshop)",
    month = mar,
    year = "2026",
    address = "Rabat, Morocco",
    publisher = "Association for Computational Linguistics",
    url = "https://aclanthology.org/2026.eacl-srw.30/",
    doi = "10.18653/v1/2026.eacl-srw.30",
    pages = "416--425",
    ISBN = "979-8-89176-383-8"
}

@inproceedings{bowman-etal-2015-large,
    title = "A large annotated corpus for learning natural language inference",
    author = "Bowman, Samuel R.  and
      Angeli, Gabor  and
      Potts, Christopher  and
      Manning, Christopher D.",
    editor = "M{\`a}rquez, Llu{\'i}s  and
      Callison-Burch, Chris  and
      Su, Jian",
    booktitle = "Proceedings of the 2015 Conference on Empirical Methods in Natural Language Processing",
    month = sep,
    year = "2015",
    address = "Lisbon, Portugal",
    publisher = "Association for Computational Linguistics",
    url = "https://aclanthology.org/D15-1075/",
    doi = "10.18653/v1/D15-1075",
    pages = "632--642"
}

@inproceedings{williams-etal-2018-broad,
    title = "A Broad-Coverage Challenge Corpus for Sentence Understanding through Inference",
    author = "Williams, Adina  and
      Nangia, Nikita  and
      Bowman, Samuel R.",
    editor = "Walker, Marilyn  and
      Ji, Heng  and
      Stent, Amanda",
    booktitle = "Proceedings of the 2018 Conference of the North {A}merican Chapter of the Association for Computational Linguistics: Human Language Technologies, Volume 1 (Long Papers)",
    month = jun,
    year = "2018",
    address = "New Orleans, Louisiana",
    publisher = "Association for Computational Linguistics",
    url = "https://aclanthology.org/N18-1101/",
    doi = "10.18653/v1/N18-1101",
    pages = "1112--1122"
}

@inproceedings{kamoi-etal-2023-wice,
    title = "{W}i{CE}: Real-World Entailment for Claims in {W}ikipedia",
    author = "Kamoi, Ryo  and
      Goyal, Tanya  and
      Diego Rodriguez, Juan  and
      Durrett, Greg",
    editor = "Bouamor, Houda  and
      Pino, Juan  and
      Bali, Kalika",
    booktitle = "Proceedings of the 2023 Conference on Empirical Methods in Natural Language Processing",
    month = dec,
    year = "2023",
    address = "Singapore",
    publisher = "Association for Computational Linguistics",
    url = "https://aclanthology.org/2023.emnlp-main.470/",
    doi = "10.18653/v1/2023.emnlp-main.470",
    pages = "7561--7583"
}

@inproceedings{romanov-shivade-2018-lessons,
    title = "Lessons from Natural Language Inference in the Clinical Domain",
    author = "Romanov, Alexey  and
      Shivade, Chaitanya",
    editor = "Riloff, Ellen  and
      Chiang, David  and
      Hockenmaier, Julia  and
      Tsujii, Jun{'}ichi",
    booktitle = "Proceedings of the 2018 Conference on Empirical Methods in Natural Language Processing",
    month = oct # "-" # nov,
    year = "2018",
    address = "Brussels, Belgium",
    publisher = "Association for Computational Linguistics",
    url = "https://aclanthology.org/D18-1187/",
    doi = "10.18653/v1/D18-1187",
    pages = "1586--1596"
}

@inproceedings{herlihy-rudinger-2021-mednli,
    title = "{M}ed{NLI} Is Not Immune: {N}atural Language Inference Artifacts in the Clinical Domain",
    author = "Herlihy, Christine  and
      Rudinger, Rachel",
    editor = "Zong, Chengqing  and
      Xia, Fei  and
      Li, Wenjie  and
      Navigli, Roberto",
    booktitle = "Proceedings of the 59th Annual Meeting of the Association for Computational Linguistics and the 11th International Joint Conference on Natural Language Processing (Volume 2: Short Papers)",
    month = aug,
    year = "2021",
    address = "Online",
    publisher = "Association for Computational Linguistics",
    url = "https://aclanthology.org/2021.acl-short.129/",
    doi = "10.18653/v1/2021.acl-short.129",
    pages = "1020--1027"
}

@inproceedings{jullien-etal-2024-semeval,
    title = "{S}em{E}val-2024 Task 2: Safe Biomedical Natural Language Inference for Clinical Trials",
    author = "Jullien, Mael  and
      Valentino, Marco  and
      Freitas, Andr{\'e}",
    editor = {Ojha, Atul Kr.  and
      Do{\u{g}}ru{\"o}z, A. Seza  and
      Tayyar Madabushi, Harish  and
      Da San Martino, Giovanni  and
      Rosenthal, Sara  and
      Ros{\'a}, Aiala},
    booktitle = "Proceedings of the 18th International Workshop on Semantic Evaluation (SemEval-2024)",
    month = jun,
    year = "2024",
    address = "Mexico City, Mexico",
    publisher = "Association for Computational Linguistics",
    url = "https://aclanthology.org/2024.semeval-1.271/",
    doi = "10.18653/v1/2024.semeval-1.271",
    pages = "1947--1962"
}

@inproceedings{mor-lan-levi-2024-exploring,
    title = "Exploring Factual Entailment with {NLI}: A News Media Study",
    author = "Mor-Lan, Guy  and
      Levi, Effi",
    editor = "Bollegala, Danushka  and
      Shwartz, Vered",
    booktitle = "Proceedings of the 13th Joint Conference on Lexical and Computational Semantics (*SEM 2024)",
    month = jun,
    year = "2024",
    address = "Mexico City, Mexico",
    publisher = "Association for Computational Linguistics",
    url = "https://aclanthology.org/2024.starsem-1.15/",
    doi = "10.18653/v1/2024.starsem-1.15",
    pages = "190--199"
}

@InProceedings{munnangi2025factehrdatasetevaluatingfactuality,
  title = 	 {Fact{EHR}: A Dataset for Evaluating Factuality in Clinical Notes Using {LLM}s},
  author =       {Munnangi, Monica and Swaminathan, Akshay and Fries, Jason Alan and Jindal, Jenelle A and Narayanan, Sanjana and Lopez, Ivan and Tu, Lucia and Chung, Philip and Omiye, Jesutofunmi and Kashyap, Mehr and Shah, Nigam},
  booktitle = 	 {Proceedings of the 10th Machine Learning for Healthcare Conference},
  year = 	 {2025},
  editor = 	 {Agrawal, Monica and Deshpande, Kaivalya and Engelhard, Matthew and Joshi, Shalmali and Tang, Shengpu and Urteaga, Iñigo},
  volume = 	 {298},
  series = 	 {Proceedings of Machine Learning Research},
  month = 	 {15--16 Aug},
  publisher =    {PMLR},
  url = 	 {https://proceedings.mlr.press/v298/munnangi25a.html}
}

@article{fitzpatrick2023improving,
  title={Improving health literacy using the power of digital communications to achieve better health outcomes for patients and practitioners},
  author={Fitzpatrick, Patrick J},
  journal={Frontiers in digital health},
  volume={5},
  pages={1264780},
  year={2023},
  publisher={Frontiers}
}

@inproceedings{maynez-etal-2020-faithfulness,
    title = "On Faithfulness and Factuality in Abstractive Summarization",
    author = "Maynez, Joshua  and
      Narayan, Shashi  and
      Bohnet, Bernd  and
      McDonald, Ryan",
    editor = "Jurafsky, Dan  and
      Chai, Joyce  and
      Schluter, Natalie  and
      Tetreault, Joel",
    booktitle = "Proceedings of the 58th Annual Meeting of the Association for Computational Linguistics",
    month = jul,
    year = "2020",
    address = "Online",
    publisher = "Association for Computational Linguistics",
    url = "https://aclanthology.org/2020.acl-main.173/",
    doi = "10.18653/v1/2020.acl-main.173",
    pages = "1906--1919"
}

@inproceedings{
clark2025epistemic,
title={Epistemic Alignment: A Mediating Framework for User-{LLM} Knowledge Delivery},
author={Nicholas Clark and Hua Shen and Bill Howe and Tanu Mitra},
booktitle={Second Conference on Language Modeling},
year={2025},
url={https://openreview.net/forum?id=Orvjm9UqH2}
}

@misc{zhu2026causalstoriessensortraces,
      title={Causal Stories from Sensor Traces: Auditing Epistemic Overreach in LLM-Generated Personal Sensing Explanations}, 
      author={Shanshan Zhu and Han Zhang and J. Doris Chi and Subigya Nepal and Koustuv Saha},
      year={2026},
      eprint={2605.08590},
      archivePrefix={arXiv},
      primaryClass={cs.HC},
      url={https://arxiv.org/abs/2605.08590}, 
}

@inproceedings{oukelmoun-etal-2025-detecting,
    title = "Detecting Omissions in {LLM}-Generated Medical Summaries",
    author = {Oukelmoun, Achir  and
      Semmar, Nasredine  and
      de Chalendar, Ga{\"e}l  and
      Cormi, Clement  and
      Oukelmoun, Mariame  and
      Vibert, Eric  and
      Allard, Marc-Antoine},
    editor = "Potdar, Saloni  and
      Rojas-Barahona, Lina  and
      Montella, Sebastien",
    booktitle = "Proceedings of the 2025 Conference on Empirical Methods in Natural Language Processing: Industry Track",
    month = nov,
    year = "2025",
    address = "Suzhou (China)",
    publisher = "Association for Computational Linguistics",
    url = "https://aclanthology.org/2025.emnlp-industry.22/",
    doi = "10.18653/v1/2025.emnlp-industry.22",
    pages = "325--337",
    ISBN = "979-8-89176-333-3"
}

@article{bull2019real,
  title={Real-world menstrual cycle characteristics of more than 600,000 menstrual cycles},
  author={Bull, Jonathan R and Rowland, Simon P and Scherwitzl, Elina Berglund and Scherwitzl, Raoul and Danielsson, Kristina Gemzell and Harper, Joyce},
  journal={NPJ digital medicine},
  volume={2},
  number={1},
  pages={83},
  year={2019},
  publisher={Nature Publishing Group UK London}
}

@article{fehring2006variability,
author = {Fehring, Richard J. and Schneider, Mary and Raviele, Kathleen},
title = {Variability in the Phases of the Menstrual Cycle},
journal = {Journal of Obstetric, Gynecologic, \& Neonatal Nursing},
volume = {35},
number = {3},
pages = {376-384},
doi = {https://doi.org/10.1111/j.1552-6909.2006.00051.x},
url = {https://onlinelibrary.wiley.com/doi/abs/10.1111/j.1552-6909.2006.00051.x},
eprint = {https://onlinelibrary.wiley.com/doi/pdf/10.1111/j.1552-6909.2006.00051.x},
year = {2006}
}

@article{baker2007sleep,
  title={Circadian rhythms, sleep, and the menstrual cycle},
  author={Baker, Fiona C and Driver, Helen S},
  journal={Sleep medicine},
  volume={8},
  number={6},
  pages={613--622},
  year={2007},
  publisher={Elsevier}
}

@article{shilaih2018bbt,
  title={Modern fertility awareness methods: wrist wearables capture the changes in temperature associated with the menstrual cycle},
  author={Shilaih, Mohaned and Goodale, Brianna M and Falco, Lisa and K{\"u}bler, Florian and De Clerck, Valerie and Leeners, Brigitte},
  journal={Bioscience reports},
  volume={38},
  number={6},
  pages={BSR20171279},
  year={2018},
  publisher={Portland Press Ltd.}
}

@incollection{wilcoxon1992individual,
  title={Individual comparisons by ranking methods},
  author={Wilcoxon, Frank},
  booktitle={Breakthroughs in statistics: Methodology and distribution},
  pages={196--202},
  year={1992},
  publisher={Springer}
}

@techreport{anthropic2024haikuaddendum,
  author      = {Anthropic},
  title       = {Model Card Addendum: Claude 3.5 Haiku and Upgraded Claude 3.5 Sonnet},
  institution = {Anthropic},
  year        = {2024},
  url         = {https://assets.anthropic.com/m/1cd9d098ac3e6467/original/Claude-3-Model-Card-October-Addendum.pdf}
}

@misc{openai2024gpt4technicalreport,
      title={GPT-4 Technical Report}, 
      author={OpenAI and Josh Achiam and Steven Adler and Sandhini Agarwal and Lama Ahmad and Ilge Akkaya and Florencia Leoni Aleman and Diogo Almeida and Janko Altenschmidt and Sam Altman and Shyamal Anadkat and Red Avila and Igor Babuschkin and Suchir Balaji and Valerie Balcom and Paul Baltescu and Haiming Bao and Mohammad Bavarian and Jeff Belgum and Irwan Bello and Jake Berdine and Gabriel Bernadett-Shapiro and Christopher Berner and Lenny Bogdonoff and Oleg Boiko and Madelaine Boyd and Anna-Luisa Brakman and Greg Brockman and Tim Brooks and Miles Brundage and Kevin Button and Trevor Cai and Rosie Campbell and Andrew Cann and Brittany Carey and Chelsea Carlson and Rory Carmichael and Brooke Chan and Che Chang and Fotis Chantzis and Derek Chen and Sully Chen and Ruby Chen and Jason Chen and Mark Chen and Ben Chess and Chester Cho and Casey Chu and Hyung Won Chung and Dave Cummings and Jeremiah Currier and Yunxing Dai and Cory Decareaux and Thomas Degry and Noah Deutsch and Damien Deville and Arka Dhar and David Dohan and Steve Dowling and Sheila Dunning and Adrien Ecoffet and Atty Eleti and Tyna Eloundou and David Farhi and Liam Fedus and Niko Felix and Simón Posada Fishman and Juston Forte and Isabella Fulford and Leo Gao and Elie Georges and Christian Gibson and Vik Goel and Tarun Gogineni and Gabriel Goh and Rapha Gontijo-Lopes and Jonathan Gordon and Morgan Grafstein and Scott Gray and Ryan Greene and Joshua Gross and Shixiang Shane Gu and Yufei Guo and Chris Hallacy and Jesse Han and Jeff Harris and Yuchen He and Mike Heaton and Johannes Heidecke and Chris Hesse and Alan Hickey and Wade Hickey and Peter Hoeschele and Brandon Houghton and Kenny Hsu and Shengli Hu and Xin Hu and Joost Huizinga and Shantanu Jain and Shawn Jain and Joanne Jang and Angela Jiang and Roger Jiang and Haozhun Jin and Denny Jin and Shino Jomoto and Billie Jonn and Heewoo Jun and Tomer Kaftan and Łukasz Kaiser and Ali Kamali and Ingmar Kanitscheider and Nitish Shirish Keskar and Tabarak Khan and Logan Kilpatrick and Jong Wook Kim and Christina Kim and Yongjik Kim and Jan Hendrik Kirchner and Jamie Kiros and Matt Knight and Daniel Kokotajlo and Łukasz Kondraciuk and Andrew Kondrich and Aris Konstantinidis and Kyle Kosic and Gretchen Krueger and Vishal Kuo and Michael Lampe and Ikai Lan and Teddy Lee and Jan Leike and Jade Leung and Daniel Levy and Chak Ming Li and Rachel Lim and Molly Lin and Stephanie Lin and Mateusz Litwin and Theresa Lopez and Ryan Lowe and Patricia Lue and Anna Makanju and Kim Malfacini and Sam Manning and Todor Markov and Yaniv Markovski and Bianca Martin and Katie Mayer and Andrew Mayne and Bob McGrew and Scott Mayer McKinney and Christine McLeavey and Paul McMillan and Jake McNeil and David Medina and Aalok Mehta and Jacob Menick and Luke Metz and Andrey Mishchenko and Pamela Mishkin and Vinnie Monaco and Evan Morikawa and Daniel Mossing and Tong Mu and Mira Murati and Oleg Murk and David Mély and Ashvin Nair and Reiichiro Nakano and Rajeev Nayak and Arvind Neelakantan and Richard Ngo and Hyeonwoo Noh and Long Ouyang and Cullen O'Keefe and Jakub Pachocki and Alex Paino and Joe Palermo and Ashley Pantuliano and Giambattista Parascandolo and Joel Parish and Emy Parparita and Alex Passos and Mikhail Pavlov and Andrew Peng and Adam Perelman and Filipe de Avila Belbute Peres and Michael Petrov and Henrique Ponde de Oliveira Pinto and Michael and Pokorny and Michelle Pokrass and Vitchyr H. Pong and Tolly Powell and Alethea Power and Boris Power and Elizabeth Proehl and Raul Puri and Alec Radford and Jack Rae and Aditya Ramesh and Cameron Raymond and Francis Real and Kendra Rimbach and Carl Ross and Bob Rotsted and Henri Roussez and Nick Ryder and Mario Saltarelli and Ted Sanders and Shibani Santurkar and Girish Sastry and Heather Schmidt and David Schnurr and John Schulman and Daniel Selsam and Kyla Sheppard and Toki Sherbakov and Jessica Shieh and Sarah Shoker and Pranav Shyam and Szymon Sidor and Eric Sigler and Maddie Simens and Jordan Sitkin and Katarina Slama and Ian Sohl and Benjamin Sokolowsky and Yang Song and Natalie Staudacher and Felipe Petroski Such and Natalie Summers and Ilya Sutskever and Jie Tang and Nikolas Tezak and Madeleine B. Thompson and Phil Tillet and Amin Tootoonchian and Elizabeth Tseng and Preston Tuggle and Nick Turley and Jerry Tworek and Juan Felipe Cerón Uribe and Andrea Vallone and Arun Vijayvergiya and Chelsea Voss and Carroll Wainwright and Justin Jay Wang and Alvin Wang and Ben Wang and Jonathan Ward and Jason Wei and CJ Weinmann and Akila Welihinda and Peter Welinder and Jiayi Weng and Lilian Weng and Matt Wiethoff and Dave Willner and Clemens Winter and Samuel Wolrich and Hannah Wong and Lauren Workman and Sherwin Wu and Jeff Wu and Michael Wu and Kai Xiao and Tao Xu and Sarah Yoo and Kevin Yu and Qiming Yuan and Wojciech Zaremba and Rowan Zellers and Chong Zhang and Marvin Zhang and Shengjia Zhao and Tianhao Zheng and Juntang Zhuang and William Zhuk and Barret Zoph},
      year={2024},
      eprint={2303.08774},
      archivePrefix={arXiv},
      primaryClass={cs.CL},
      url={https://arxiv.org/abs/2303.08774}, 
}

@misc{qwen2025qwen25technicalreport,
      title={Qwen2.5 Technical Report}, 
      author={{Qwen} and : and An Yang and Baosong Yang and Beichen Zhang and Binyuan Hui and Bo Zheng and Bowen Yu and Chengyuan Li and Dayiheng Liu and Fei Huang and Haoran Wei and Huan Lin and Jian Yang and Jianhong Tu and Jianwei Zhang and Jianxin Yang and Jiaxi Yang and Jingren Zhou and Junyang Lin and Kai Dang and Keming Lu and Keqin Bao and Kexin Yang and Le Yu and Mei Li and Mingfeng Xue and Pei Zhang and Qin Zhu and Rui Men and Runji Lin and Tianhao Li and Tianyi Tang and Tingyu Xia and Xingzhang Ren and Xuancheng Ren and Yang Fan and Yang Su and Yichang Zhang and Yu Wan and Yuqiong Liu and Zeyu Cui and Zhenru Zhang and Zihan Qiu},
      year={2025},
      eprint={2412.15115},
      archivePrefix={arXiv},
      primaryClass={cs.CL},
      url={https://arxiv.org/abs/2412.15115}, 
}

@misc{microsoft2025phi4minitechnicalreportcompact,
      title={Phi-4-Mini Technical Report: Compact yet Powerful Multimodal Language Models via Mixture-of-LoRAs}, 
      author={{Microsoft} and : and Abdelrahman Abouelenin and Atabak Ashfaq and Adam Atkinson and Hany Awadalla and Nguyen Bach and Jianmin Bao and Alon Benhaim and Martin Cai and Vishrav Chaudhary and Congcong Chen and Dong Chen and Dongdong Chen and Junkun Chen and Weizhu Chen and Yen-Chun Chen and Yi-ling Chen and Qi Dai and Xiyang Dai and Ruchao Fan and Mei Gao and Min Gao and Amit Garg and Abhishek Goswami and Junheng Hao and Amr Hendy and Yuxuan Hu and Xin Jin and Mahmoud Khademi and Dongwoo Kim and Young Jin Kim and Gina Lee and Jinyu Li and Yunsheng Li and Chen Liang and Xihui Lin and Zeqi Lin and Mengchen Liu and Yang Liu and Gilsinia Lopez and Chong Luo and Piyush Madan and Vadim Mazalov and Arindam Mitra and Ali Mousavi and Anh Nguyen and Jing Pan and Daniel Perez-Becker and Jacob Platin and Thomas Portet and Kai Qiu and Bo Ren and Liliang Ren and Sambuddha Roy and Ning Shang and Yelong Shen and Saksham Singhal and Subhojit Som and Xia Song and Tetyana Sych and Praneetha Vaddamanu and Shuohang Wang and Yiming Wang and Zhenghao Wang and Haibin Wu and Haoran Xu and Weijian Xu and Yifan Yang and Ziyi Yang and Donghan Yu and Ishmam Zabir and Jianwen Zhang and Li Lyna Zhang and Yunan Zhang and Xiren Zhou},
      year={2025},
      eprint={2503.01743},
      archivePrefix={arXiv},
      primaryClass={cs.CL},
      url={https://arxiv.org/abs/2503.01743}, 
}

@inproceedings{deka-etal-2023-multiple,
    title = "Multiple Evidence Combination for Fact-Checking of Health-Related Information",
    author = "Deka, Pritam  and
      Jurek-Loughrey, Anna  and
      P, Deepak",
    booktitle = "The 22nd Workshop on Biomedical Natural Language Processing and BioNLP Shared Tasks",
    month = jul,
    year = "2023",
    address = "Toronto, Canada",
    publisher = "Association for Computational Linguistics",
    url = "https://aclanthology.org/2023.bionlp-1.20",
    pages = "237--247",
}

@article{10.5665/sleep.6218,
    author = {Castro-Diehl, Cecilia and Diez Roux, Ana V and Redline, Susan and Seeman, Teresa and McKinley, Paula and Sloan, Richard and Shea, Steven},
    title = {Sleep Duration and Quality in Relation to Autonomic Nervous System Measures: The Multi-Ethnic Study of Atherosclerosis (MESA)},
    journal = {Sleep},
    volume = {39},
    number = {11},
    pages = {1927-1940},
    year = {2016},
    month = {11},
    issn = {0161-8105},
    doi = {10.5665/sleep.6218},
    url = {https://doi.org/10.5665/sleep.6218},
    eprint = {https://academic.oup.com/sleep/article-pdf/39/11/1927/28832081/aasm.39.11.1927.pdf},
}

@article{https://doi.org/10.1111/psyp.12250,
author = {Tenan, Matthew S. and Brothers, R. Matthew and Tweedell, Andrew J. and Hackney, Anthony C. and Griffin, Lisa},
title = {Changes in resting heart rate variability across the menstrual cycle},
journal = {Psychophysiology},
volume = {51},
number = {10},
pages = {996-1004},
doi = {https://doi.org/10.1111/psyp.12250},
url = {https://onlinelibrary.wiley.com/doi/abs/10.1111/psyp.12250},
eprint = {https://onlinelibrary.wiley.com/doi/pdf/10.1111/psyp.12250},
year = {2014}
}

\appendix
\section{Synthetic Data Generation: Causal DAG Specification}
\label{app:datagen}

This appendix details the generative model used to produce the semi-synthetic menstrual health dataset described in Section~\ref{sec:dataset}. The generator implements a timeline-first causal DAG in which each variable is sampled conditional only on its causal parents, ensuring that inter-variable correlations arise from shared physiological mechanisms rather than arbitrary co-occurrence.

\subsection{Causal Graph Structure}

The generation follows a strict topological ordering:

\begin{quote}\small
\textit{Cycle Length} $\rightarrow$ \textit{Luteal Length} $\rightarrow$ \textit{Phase Boundaries} $\rightarrow$ \textit{Cycle Day} $\rightarrow$ \textit{Phase} 

$\quad\hookrightarrow$ \textit{Sleep Quality} $\rightarrow$ \textit{Sleep Duration}

$\quad\hookrightarrow$ \textit{RHR, HRV} \quad (\textit{conditioned on Phase} $+$ \textit{Sleep})

$\quad\hookrightarrow$ $\text{SpO}_2$ \quad (\textit{independent})

$\quad\hookrightarrow$ \textit{Temperature Trend} $\rightarrow$ \textit{Symptoms}
\end{quote}

\subsection{Node 1: Cycle Length}

Cycle length is drawn from a lognormal distribution, reflecting the right-skewed population distribution documented in \citet{bull2019real} ($N > 600{,}000$ cycles):

\begin{equation}
\begin{aligned}
L_{\text{cycle}} &\sim \text{Lognormal}(\mu = 3.355,\; \sigma = 0.155), \notag \\
&\quad L_{\text{cycle}} \in [21, 50]
\end{aligned}
\end{equation}

This parameterization yields a median of approximately 28.7 days, a mean of approximately 29.3 days, and standard deviation of approximately 4.4 days.

\subsection{Node 2: Luteal Phase Length}

The luteal phase is sampled from a tight normal distribution, encoding the well-established finding that inter-cycle variability is attributable almost entirely to the follicular phase, with luteal length remaining approximately constant \cite{fehring2006variability}:

\begin{equation*}
L_{\text{luteal}} \sim \mathcal{N}(\mu = 14.0,\; \sigma = 1.5), \quad L_{\text{luteal}} \in [10, 17]
\end{equation*}

\subsection{Node 3: Phase Boundary Derivation}

Phase boundaries are computed deterministically rather than assigned via hard-coded day ranges:

\begin{align}
d_{\text{ovulation}} &= L_{\text{cycle}} - L_{\text{luteal}} \notag \\
\text{Menstrual:} &\quad \text{days } 1 - 5 \notag \\
\text{Follicular:} &\quad \text{days } 6 - (d_{\text{ovulation}} - 1) \notag \\
\text{Ovulatory:} &\quad \text{day } d_{\text{ovulation}} \notag \\
\text{Luteal:} &\quad \text{days } (d_{\text{ovulation}} + 1) - L_{\text{cycle}} \notag
\end{align}

A cycle day is then sampled uniformly: $d \sim \text{Uniform}\{1, \ldots, L_{\text{cycle}}\}$.
To keep a non-empty follicular phase, $d_{\text{ovulation}}$ is floored at day~7.

\subsection{Node 4--5: Sleep (Phase-Conditioned)}

Sleep quality is drawn from a categorical distribution conditioned on the current phase, reflecting progesterone-mediated sleep architecture disruption \cite{baker2007sleep}:

\begin{equation*}
\text{SleepQuality} \sim \text{Categorical}(P(\text{quality} \mid \text{phase}))
\end{equation*}

\begin{table}[h]
\centering
\small
\caption{Conditional probability table: $P(\text{sleep quality} \mid \text{phase})$}
\label{tab:sleep_quality}
\begin{tabular}{lccc}
\toprule
\textbf{Phase} & \textbf{Poor} & \textbf{Moderate} & \textbf{Good} \\
\midrule
Menstrual  & 0.35 & 0.40 & 0.25 \\
Follicular & 0.08 & 0.32 & 0.60 \\
Ovulatory  & 0.12 & 0.40 & 0.48 \\
Luteal     & 0.32 & 0.43 & 0.25 \\
\bottomrule
\end{tabular}
\end{table}

Sleep duration is then sampled conditional on quality from a normal distribution clipped to quality-specific bounds:

\begin{equation*}
\text{Duration} \sim \mathcal{N}(\mu_q, \sigma_q), \quad \text{clipped to } [\text{min}_q, \text{max}_q]
\end{equation*}

where the parameters $(\mu_q, \sigma_q, \text{min}_q, \text{max}_q)$ are $(4.5, 0.6, 3.5, 5.5)$ for poor, $(6.3, 0.5, 5.5, 7.2)$ for moderate, and $(7.8, 0.6, 7.0, 9.2)$ for good quality.

\subsection{Nodes 6--7: Autonomic Biometrics (Phase + Sleep Override)}

Resting heart rate is drawn from a phase-specific baseline with an additive penalty for poor sleep:

\begin{multline*}
\text{RHR} = \underbrace{\mathcal{N}(\mu_{\text{phase}}, \sigma_{\text{phase}})}_{\text{hormonal baseline}} \\
+ \underbrace{\text{Uniform}(4, 8) \cdot \mathbf{1}[\text{sleep} = \text{poor}]}_{\text{acute sympathetic penalty}}
\end{multline*}

This encodes both chronic luteal-phase elevation \cite{https://doi.org/10.1111/psyp.12250} and acute sleep-deprivation effects on sympathovagal balance \cite{10.5665/sleep.6218}.

\begin{table}[h]
\centering
\small
\caption{RHR baseline parameters by phase (bpm)}
\label{tab:rhr}
\begin{tabular}{lcccc}
\toprule
\textbf{Phase} & $\mu$ & $\sigma$ & \textbf{Min} & \textbf{Max} \\
\midrule
Menstrual  & 63.0 & 3.0 & 57 & 71 \\
Follicular & 60.0 & 2.5 & 55 & 66 \\
Ovulatory  & 64.0 & 3.0 & 58 & 72 \\
Luteal     & 67.0 & 3.5 & 61 & 78 \\
\bottomrule
\end{tabular}
\end{table}

HRV trend is sampled categorically from phase-specific probabilities, with a hard override when sleep quality is poor (forcing sympathetic-dominant states):

\begin{equation*}
P(\text{HRV trend} \mid \text{poor sleep}) = \begin{cases} 0.55 & \text{decreasing} \\ 0.45 & \text{erratic} \end{cases}
\end{equation*}
Otherwise, HRV trend follows Table~\ref{tab:hrv}, with higher vagal tone before ovulation and sympathetic dominance after it.

\begin{table}[h]
\centering
\small
\caption{Conditional probability table: $P(\text{HRV trend} \mid \text{phase})$ when sleep quality is not poor}
\label{tab:hrv}
\begin{tabular}{lcccc}
\toprule
\textbf{Phase} & \textbf{Incr.} & \textbf{Stable} & \textbf{Decr.} & \textbf{Erratic} \\
\midrule
Menstrual  & 0.45 & 0.40 & 0.10 & 0.05 \\
Follicular & 0.50 & 0.40 & 0.05 & 0.05 \\
Ovulatory  & 0.10 & 0.15 & 0.45 & 0.30 \\
Luteal     & 0.05 & 0.25 & 0.50 & 0.20 \\
\bottomrule
\end{tabular}
\end{table}

\subsection{Node 8: SpO\textsubscript{2} (Independent)}

Blood oxygen saturation is sampled independently of all other nodes, reflecting tight chemoreceptor-mediated homeostatic regulation in healthy individuals:

\begin{equation*}
\text{SpO}_2 \sim \mathcal{N}(97.8, 0.8), \quad \text{SpO}_2 \in [95.0, 100.0]
\end{equation*}

\subsection{Node 9: Temperature Trend}

Temperature trend is sampled categorically from phase-specific probabilities, encoding the well-documented biphasic thermal shift driven by post-ovulatory progesterone thermogenesis \cite{shilaih2018bbt}.
The probabilities are given in Table~\ref{tab:temp}.

\begin{table}[h]
\centering
\small
\caption{Conditional probability table: $P(\text{temperature trend} \mid \text{phase})$}
\label{tab:temp}
\begin{tabular}{lcccc}
\toprule
\textbf{Phase} & \textbf{Declin.} & \textbf{Low st.} & \textbf{Elev.} & \textbf{Biph.} \\
\midrule
Menstrual  & 0.55 & 0.40 & 0.03 & 0.02 \\
Follicular & 0.05 & 0.85 & 0.05 & 0.05 \\
Ovulatory  & 0.02 & 0.08 & 0.20 & 0.70 \\
Luteal     & 0.03 & 0.02 & 0.80 & 0.15 \\
\bottomrule
\end{tabular}
\end{table}

\subsection{Node 10: Symptoms (Causal Placeholder)}

Symptoms are generated as logical consequences of upstream states rather than as independent draws. The causal triggers include:

\begin{itemize}
    \item \textbf{Fatigue} ($p{=}0.75$): triggered by poor sleep OR elevated RHR ($> 70$ bpm)
    \item \textbf{Headache} ($p{=}0.50$): triggered by poor sleep AND declining/erratic HRV
    \item \textbf{Cramps} (0.70), \textbf{back pain} (0.35): phase-gated (menstrual)
    \item \textbf{Bloating} (0.45), \textbf{breast tenderness} (0.40), \textbf{mood changes} (0.30): phase-gated (luteal)
    \item \textbf{Bloating} (0.25), \textbf{nausea} (0.15): phase-gated (ovulatory)
    \item \textbf{Mood changes} ($p{=}0.40$): also triggered by autonomic stress (poor sleep + declining HRV)
\end{itemize}

Records with no triggered symptom are labeled \textit{none}.
This ensures that symptom co-occurrence patterns reflect plausible physiological mechanisms rather than arbitrary correlation.

\subsection{User Profile Masking}

After complete record generation, a masking layer simulates five distinct user archetypes that determine field observability:

\begin{table}[h]
\centering
\small
\caption{User profile distribution and data access characteristics}
\label{tab:profiles}
\begin{tabular}{lcc}
\toprule
\textbf{Profile} & \textbf{$N$} & \textbf{Mean Fields} \\
\midrule
Minimal tracker      & 75  & $\sim$2--3 \\
App logger           & 100 & $\sim$4--5 \\
Basic wearable       & 125 & $\sim$6--8 \\
Advanced wearable    & 125 & $\sim$9--12 \\
Inconsistent user    & 75  & $\sim$3--5 \\
\bottomrule
\end{tabular}
\end{table}

Each profile specifies fields as \textit{always present}, \textit{probabilistically available} (with per-field inclusion probabilities), or \textit{never observed}. This produces structured missingness patterns that simulate the heterogeneous data completeness characteristic of real-world self-tracking behavior.

\subsection{Confidence Score}

A data-driven confidence score is computed from the ratio of present diagnostic fields to total possible fields, modulated by phase clarity (menstrual phase being most identifiable due to bleeding, follicular least identifiable due to ambiguous signals):

\begin{equation*}
\begin{split}
C = &\ 0.30 + 0.50 \cdot \frac{|\text{present diagnostic fields}|}{|\text{total diagnostic fields}|} \\
&{}+ 0.20 \cdot \text{clarity}(\text{phase}) + \epsilon
\end{split}
\end{equation*}
where $\epsilon \sim \mathcal{N}(0, 0.03)$ and clarity values are $\{$menstrual: 0.90, luteal: 0.80, ovulatory: 0.75, follicular: 0.70$\}$.

\subsection{Causal Coherence Validation}

To verify that the DAG produces internally consistent data, we report two key checks on the generated dataset:

\begin{itemize}
    \item \textbf{Sleep $\rightarrow$ RHR}: Mean RHR under poor sleep is elevated by $+$6--8 bpm compared to good sleep, consistent with the encoded sympathetic penalty.
    \item \textbf{Sleep $\rightarrow$ HRV override}: When sleep is poor, 100\% of HRV trends are either ``decreasing'' or ``erratic,'' confirming the hard override mechanism.
\end{itemize}

\subsection{Reproducibility}

All stochastic operations are seeded (seed = 42). The generation script and output dataset are available at   \href{https://github.com/Beyramayadi/clinical-alignment-public}{https://github.com/Beyramayadi/clinical-alignment-public}. Total generation: 500 observed samples + 500 ground-truth records.

\subsection{Example Record}
\label{app:example-record}
An example ground-truth record produced by the generator, prior to user-profile masking, is shown below.

\begin{small}
\begin{verbatim}
{
  "cycle_day": 22,
  "phase": "luteal",
  "temperature_trend": "elevated",
  "hrv_trend": "decreasing",
  "symptoms": ["fatigue", "cramps"],
  "sleep_quality": "poor",
  "resting_hr_bpm": 71.3,
  "confidence_score": 0.81,
  "missing_fields": ["spo2_pct",
     "sleep_duration_hrs"]
}
\end{verbatim}
\end{small}
\section{System Prompts and Hyperparameters}
\label{app:system_prompts}

This appendix provides the exact system prompts and generation configurations used across the three experimental settings described in Section~\ref{sec:experimental_setup}. For all generation runs, the model temperature was set to $T = 0.4$ to balance deterministic fidelity with linguistic fluency.

\subsection{Base (Vanilla LLM) Setting}
\label{app:base_setting}

\noindent\textbf{System Prompt:}
\begin{quote}\small
\textit{``You are a helpful health assistant. Summarize the following user tracking data into a friendly, patient-facing health summary using simple language.''}
\end{quote}

\noindent\textbf{User Prompt Template:}
\begin{quote}\small
USER DATA:\\
\texttt{\{user\_data\}}

\vspace{0.5em}
Return only the summary text.
\end{quote}

\subsection{Constrained Setting}
\label{app:constrained_setting}

\noindent\textbf{System Prompt:}
\begin{quote}\small
\textit{``You are a highly precise clinical summarization assistant. Your task is to translate structured, longitudinal physiological data into an accessible, lay-language patient narrative. You must strictly follow the constraints below:}

\vspace{0.3em}
\textbf{INPUT DATA}:\\
- Structured Physiological Record ($X$): \texttt{\{user\_{}data\}}

\vspace{0.3em}
\textbf{STAGE-GATED GENERATION CONSTRAINTS}:\\
1. \textbf{Evidence Consistency}: Only write claims directly supported by the provided record. Do not mention, assume, or infer missing metrics.\\
2. \textbf{Uncertainty Preservation}: Use hedged language when confidence is low or evidence is incomplete.\\
3. \textbf{Diagnostic Restraint}: Do not infer or suggest any diagnosis, disease, or pathology.\\
4. \textbf{Nuance Preservation}: Keep important qualifiers like phase context and explicit trends.\\
5. \textbf{Tone Calibration}: Keep tone supportive and balanced; avoid panic or over-reassurance.

\vspace{0.3em}
\textbf{OUTPUT FORMAT}:\\
Return only a patient-facing summary with:\\
- Your Current Status\\
- What your data is showing\textit{''}
\end{quote}

\subsection{RAG-Constrained (Proposed) Setting}
\label{app:rag_setting}

\noindent\textbf{System Prompt:}
\begin{quote}\small
\textit{``You are a highly precise clinical summarization assistant. Your task is to translate structured, longitudinal physiological data into an accessible, lay-language patient narrative. You must strictly follow the constraints below:}

\vspace{0.3em}
\textbf{INPUT DATA}:\\
- Structured Physiological Record ($X$): \texttt{\{user\_data\}}

\vspace{0.3em}
\textbf{RETRIEVED CONTEXT}:\\
Use this context only as support for explanations. Do not contradict the record.\\
\texttt{\{retrieved\_context\}}

\vspace{0.3em}
\textbf{STAGE-GATED GENERATION CONSTRAINTS}:\\
1. \textbf{Evidence Consistency}: Only write claims directly supported by record and context.\\
2. \textbf{Uncertainty Preservation}: Use hedged language when confidence is low.\\
3. \textbf{Diagnostic Restraint}: Do not infer or suggest any diagnosis or pathology.\\
4. \textbf{Nuance Preservation}: Keep important qualifiers and explicit trends.\\
5. \textbf{Tone Calibration}: Keep tone supportive, avoiding panic/over-reassurance.

\vspace{0.3em}
\textbf{OUTPUT FORMAT}:\\
Return only a patient-facing summary with:\\
- Your Current Status\\
- What your data is showing\\
- Symptom Context\textit{''}
\end{quote}

\section{Retrieval-Augmented Generation Details}
\label{app:rag}

This appendix specifies the retrieval component used in the
RAG-Constrained setting.
Retrieval plays a deliberately narrow role: it supplies lay-term
definitions that support simplification, rather than injecting new
clinical evidence. All grounding definitions are drawn from the
README dataset \citep{yao-etal-2024-readme}, represented as
\emph{(medical term, lay definition)} pairs, where the term field is
indexed for retrieval and the definition field is returned as
grounding context. Prior to indexing, we drop rows with a missing
term or definition and remove exact duplicate pairs after case- and
whitespace-normalization, so that a single definition cannot occupy
multiple retrieval slots; only the deduplicated, non-null rows are
indexed.

We use a sparse TF--IDF retriever with cosine similarity. At this corpus size, TF--IDF is
fully local (requiring no model download or additional network
dependency), fast to build and query, and provides a transparent,
inspectable matching signal, while leaving a clear upgrade path to
dense retrieval without changing the surrounding pipeline. The term
index is built with an \texttt{sklearn} \texttt{TfidfVectorizer} using
English stop-word removal and an $n$-gram range of $(1,2)$; combining
unigram and bigram features allows multi-word terms such as
\emph{``menstrual cramps''} to match on shared unigrams while still
retaining some phrase-level signal. For a given query $q$, we compute
cosine similarity against all indexed terms, rank candidates in
descending order, and return at most $k=3$ definitions, retaining only
matches with similarity at least $0.12$. This similarity floor treats
weak lexical overlap as ``no match'' rather than forcing spurious
grounding, and any duplicate definitions surviving to the ranking
stage are skipped so that the injected context contains distinct
entries.
The concatenated text forms the retrieval query, and the top-$k$ retrieved definitions
are formatted as a bulleted \textsc{Retrieved Context} block and
inserted into the RAG-Constrained prompt (Appendix~\ref{app:system_prompts}).

The prompt further constrains how retrieved content may be used:
definitions serve only as support for explanations and wording, and
must not contradict or extend the structured record $X$.

\subsection{Knowledge Base Scope and Maintenance}
\label{app:rag_kb}
Retrieval is scoped strictly to lay-language definitions of medical terms rather than open-ended clinical retrieval, which keeps the knowledge base bounded and static for the experiments reported here. Any use beyond this study would nevertheless require maintaining it: README is a fixed snapshot, so the lexicon should be versioned and periodically updated as terminology and patient-communication guidance evolve, and new or modified entries should be reviewed by a human, ideally a clinician or health-communication expert, before being indexed. Recording the lexicon version alongside each generated summary would further allow outputs to be traced back to the definitions available at generation time.

\section{Readability metrics}
\label{sec:readability}
Table~\ref{tab:readability} reports all grade-level readability scores; FKGL and Dale-Chall are also included in Table~\ref{tab:clinical_alignment_combined}. Pairwise correlations between readability metrics are shown in Appendix~\ref{sec:correlations}.

\begin{table*}[htbp]
\centering
\small
\begin{tabular}{llccc}
\toprule
Model & Metric & Base & Constrained & RAG-constrained \\
\midrule
GPT-4o & FKGL & 7.38 & 9.93 & 10.31 \\
GPT-4o & ARI & 8.35 & 10.97 & 10.83 \\
GPT-4o & Dale-Chall & 9.67 & 10.60 & 10.54 \\
Phi-4-mini & FKGL & 10.63 & 12.18 & 12.32 \\
Phi-4-mini & ARI & 12.63 & 14.13 & 14.29 \\
Phi-4-mini & Dale-Chall & 9.84 & 10.44 & 10.38 \\
Qwen2.5-14B & FKGL & 7.64 & 9.72 & 10.39 \\
Qwen2.5-14B & ARI & 8.07 & 10.66 & 11.64 \\
Qwen2.5-14B & Dale-Chall & 10.07 & 11.01 & 10.92 \\
\bottomrule
\end{tabular}
\caption{Mean grade-level readability scores by model and generation setting.}
\label{tab:readability}
\end{table*}

\section{Medical Jargon-Related Metrics}
\label{sec:jargon}
\begin{figure*}[htbp]
  \centering
  \includegraphics[width=\textwidth]{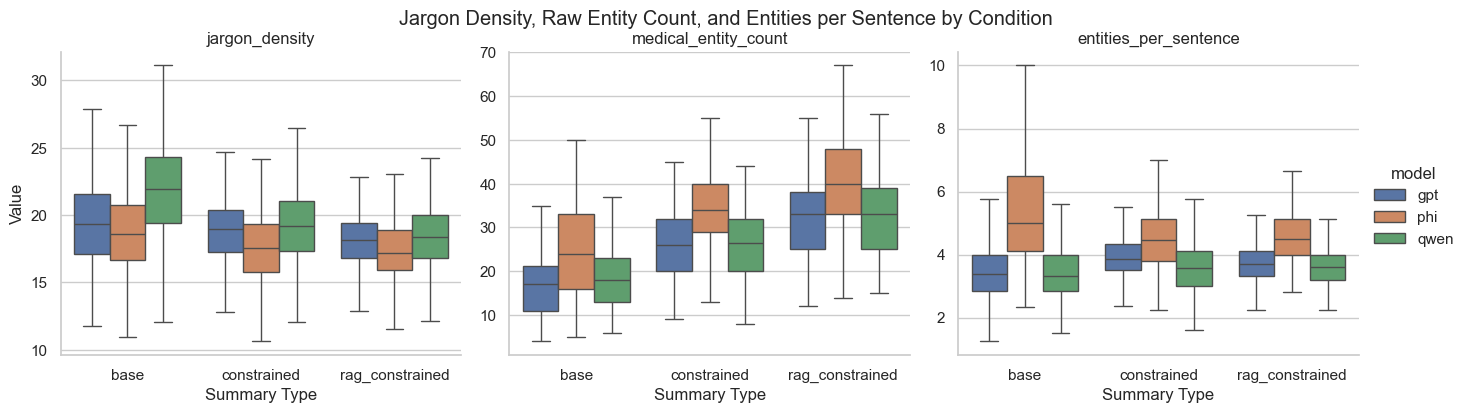}
  \caption{Distribution of jargon density, raw medical entity counts, and entities per sentence across generation conditions (base, constrained, rag\_constrained) for all evaluated models.}
  \label{fig:jargon_metrics}
\end{figure*}

Figure~\ref{fig:jargon_metrics} shows the distributions and Table~\ref{tab:jargon_metrics} the mean values of the jargon-related metrics; medical-entity counts are also included in Table~\ref{tab:clinical_alignment_combined}.

\begin{table*}[htbp]
  \centering
  \small
  \setlength{\tabcolsep}{6pt}
  \begin{tabular}{llrrrr}
    \toprule
    \textbf{Model} & \textbf{Summary Type} & \textbf{Jargon} & \textbf{Med. Ent.} & \textbf{Ent./Sent.} & \textbf{Sent.} \\
    \midrule
    GPT-4o & base & 19.26 & 16.63 & 3.51 & 4.80 \\
    Phi-4 & base & 18.91 & 24.28 & 5.59 & 4.97 \\
    Qwen & base & 21.90 & 18.38 & 3.44 & 5.34 \\
    \midrule
    GPT-4o & constrained & 18.92 & 26.07 & 3.92 & 6.67 \\
    Phi-4 & constrained & 17.65 & 33.61 & 4.54 & 7.65 \\
    Qwen & constrained & 19.31 & 26.26 & 3.63 & 7.45 \\
    \midrule
    GPT-4o & rag-constr. & 18.07 & 32.29 & 3.78 & 8.65 \\
    Phi-4 & rag-constr. & 17.37 & 40.96 & 4.57 & 9.14 \\
    Qwen & rag-constr. & 18.37 & 32.44 & 3.63 & 9.04 \\
    \bottomrule
  \end{tabular}
  \caption{Summary of medical jargon-related metrics (mean values).}
  \label{tab:jargon_metrics}
\end{table*}

\FloatBarrier
\section{Metric correlations}
\label{sec:correlations}
\begin{figure}[!htbp]
    \centering
    \includegraphics[width=0.99\linewidth]{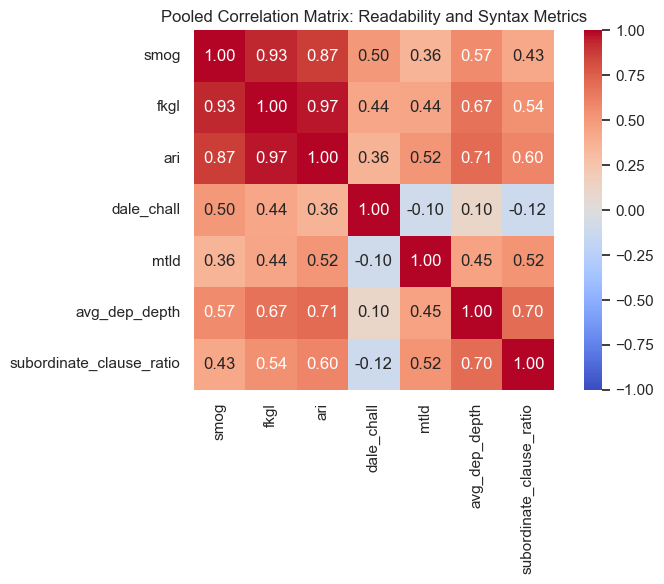}
   
    \caption{Pooled pairwise correlations ($r$) across seven readability, lexical, and syntactic-complexity metrics computed over all $4,500$ generated summaries. Grade-level indices are strongly intercorrelated ($r \ge 0.87$).}
    \label{fig:metrics_corr}
\end{figure}

Figure~\ref{fig:metrics_corr} shows pooled pairwise correlations between the readability, lexical, and syntactic-complexity metrics, which motivate using FKGL as the primary grade-level endpoint (\S\ref{sec:evaluation}).

\FloatBarrier

\section{Qualitative Examples}
\label{app:qual_examples}
Tables~\ref{tab:qual_full_gpt4o}--\ref{tab:qual_full_phi} give the complete summaries generated for record MH-0094 by each model and setting; Table~\ref{tab:qual_examples} shows excerpts of the GPT-4o summaries. The record was selected automatically, not by inspecting outputs: among records with all nine summaries, at least one missing field or a confidence score below 0.60, and a profile other than \textit{minimal\_tracker}, it has the smallest sum of squared standardized deviations from the per-model, per-setting means of word count, FKGL, and medical-entity count. The masked input is: Day 8, follicular; cycle length 30 days; sleep: good, 7.97\,h; RHR 65.19\,bpm; HRV trend: stable; SpO$_2$ 98.5\%; symptoms: none; note: ``Feeling a bit off today, probably just mentally drained [\ldots]''; confidence 0.86. Summaries are reproduced verbatim; Markdown headers, bold text, and bullets produced by the models are rendered, and hedge markers used by $M_{\mathrm{epist}}$ are in italics.

\begin{table*}[htbp]
\centering
\footnotesize
\begin{tabular}{p{0.30\textwidth}p{0.30\textwidth}p{0.30\textwidth}}
\toprule
\textbf{Base} & \textbf{Constrained} & \textbf{RAG-Constrained} \\
\midrule
You're on day 8 of your cycle, currently in the follicular phase, with a typical cycle length of about 30 days. Your sleep quality was good, and you got nearly 8 hours of rest, which is great! Your resting heart rate and oxygen levels look healthy, and your heart rate variability trend is stable. You’re not experiencing any physical symptoms, but it sounds like you’re feeling a bit mentally drained from a busy week. Consider taking some downtime and reducing screen time to recharge. Keep up the good work caring for yourself!
&
\textbf{Your Current Status} \newline Today is day 8 of your cycle, and you're in the follicular phase. You mentioned feeling a bit off, which you think \textit{might} be due to mental fatigue from a busy week. It sounds like you \textit{might} benefit from some downtime and reducing screen time to recharge. \par\smallskip \textbf{What Your Data is Showing} \newline Your sleep quality was good, and you got nearly 8 hours of rest, which is great for recovery. Your resting heart rate is at 65 beats per minute, and your heart rate variability trend \textit{appears} stable, both of which \textit{suggest} your body is maintaining balance. Your oxygen saturation is at a healthy level of 98.5\%. There are no physical symptoms reported today, and overall, your physiological data looks steady.
&
\textbf{Your Current Status} \newline You are currently on day 8 of your menstrual cycle, which places you in the follicular phase. This is the phase that occurs after your period and before ovulation. Based on your reported cycle length of 30 days, this timing is \textit{consistent with} what is typically expected. \par\smallskip \textbf{What Your Data Is Showing} \newline Your sleep quality was reported as good, and you slept for about 8 hours, which is generally considered a healthy duration. Your resting heart rate was measured at approximately 65 beats per minute, and your heart rate variability (HRV) is stable, which \textit{suggests} that your body is managing stress and recovery well at this time. Additionally, your oxygen saturation level (SpO2) was 98.5\%, which is within the normal range for most people. \par\smallskip \textbf{Symptom Context} \newline You did not report any symptoms today, which \textit{may indicate} that you are feeling well overall. The data provided has a confidence score of 0.86, which \textit{suggests} a fairly reliable snapshot of your current state. If you notice any changes or have concerns, it’s always a good idea to monitor how you’re feeling and reach out to a healthcare provider if needed. \\
\bottomrule
\end{tabular}
\caption{Full GPT-4o summaries of record MH-0094 under the three generation settings.}
\label{tab:qual_full_gpt4o}
\end{table*}

\begin{table*}[htbp]
\centering
\footnotesize
\begin{tabular}{p{0.30\textwidth}p{0.30\textwidth}p{0.30\textwidth}}
\toprule
\textbf{Base} & \textbf{Constrained} & \textbf{RAG-Constrained} \\
\midrule
Today is day 8 of your cycle in the follicular phase. You had good sleep for about 8 hours last night, which is great! Your resting heart rate was 65 BPM and your HRV trend is stable, indicating that your body is doing well overall. Oxygen saturation levels are at 98.5\%, also very healthy. No physical symptoms were reported today. However, you mentioned feeling mentally drained from managing multiple tasks this week. It \textit{might} be helpful to take some time for relaxation and reduce screen time to recharge.
&
Your current status today shows that you are feeling slightly off, \textit{possibly} due to mental fatigue from managing multiple tasks this week. Here’s what your data indicates: \par\smallskip \textbullet~You are currently in the follicular phase of your menstrual cycle on day 8. \newline \textbullet~Your sleep quality was good and you slept for approximately 8 hours last night. \newline \textbullet~Your resting heart rate is stable at around 65 beats per minute. \newline \textbullet~Oxygen saturation levels were recorded at 98.5\%, which is within a healthy range. \newline \textbullet~You reported no physical symptoms today. \par\smallskip The confidence score of your current status is moderate (0.86), suggesting that while the data generally aligns with how you feel, there \textit{might} be some variability in your condition or measurement accuracy. \par\smallskip It sounds like you \textit{could} benefit from taking some time to rest and reduce screen exposure this week to help manage your mental energy levels.
&
\textbf{Your Current Status:} \newline Today marks day 8 of your menstrual cycle. You are currently in the follicular phase, which typically lasts until ovulation occurs. This phase is characterized by the development and maturation of ovarian follicles. \par\smallskip \textbf{What Your Data Is Showing:} \newline \textbullet~\textbf{Sleep Quality:} You had a good night's sleep. \newline \textbullet~\textbf{Sleep Duration:} You slept for approximately 8 hours last night. \newline \textbullet~\textbf{Resting Heart Rate (BPM):} Your resting heart rate was measured at around 65 BPM, which is within the normal range for most adults. \newline \textbullet~\textbf{Heart Rate Variability (HRV) Trend:} Your HRV trend is stable, indicating that your body is managing stress well and recovering adequately between heartbeats. \newline \textbullet~\textbf{Oxygen Saturation (SpO2):} Your blood oxygen level was 98.5\%, which is a healthy reading. \par\smallskip \textbf{Symptom Context:} \newline You reported no symptoms today, which aligns with the typical symptom profile during this phase of your cycle. The confidence score for these observations is 0.86, indicating that the data is reliable but there's still some room for variability. \par\smallskip If you have any concerns or notice changes in how you feel, it’s always a good idea to reach out to your healthcare provider for further guidance. \\
\bottomrule
\end{tabular}
\caption{Full Qwen2.5-14B summaries of record MH-0094 under the three generation settings.}
\label{tab:qual_full_qwen}
\end{table*}

\begin{table*}[htbp]
\centering
\footnotesize
\begin{tabular}{p{0.30\textwidth}p{0.30\textwidth}p{0.30\textwidth}}
\toprule
\textbf{Base} & \textbf{Constrained} & \textbf{RAG-Constrained} \\
\midrule
Hello! Here's your health update for cycle day 8: You're in phase 2 of your menstrual cycle which typically lasts about 30 days on average – that's a good sleep duration with you resting around 7 hours each night, at an even heart rate and excellent oxygen levels (98.5\% SpO2). Your HRV is stable too! No symptoms reported today but you've mentioned feeling mentally drained; maybe it's time for some rest away from the screens to recharge your batteries fully. \par\smallskip Take care of yourself – you deserve a break sometimes!
&
Your current status during the 8th day of an average menstrual cycle (which lasts about 30 days) shows that you are \textit{likely} experiencing the follicular phase. You're sleeping well for most people—around 7 hours and 58 minutes per night—with a good quality score, which means you're getting restful sleep. \par\smallskip Today you've felt somewhat off-balance; this \textit{might} be due to feeling mentally drained from managing many tasks over the past week or \textit{possibly} needing some rest with less screen time until you feel better. Your heart rate while resting is normal at around 65 beats per minute (bpm), and your oxygen levels in blood are excellent. \par\smallskip We've noticed that these observations match closely what you'd expect for someone not experiencing any symptoms, but there's a slight uncertainty because the confidence score isn't perfect—it's quite high though! Just remember to listen to yourself: if you need some downtime or less screen time before tomorrow. Take care of it like your body is telling you now.
&
\textbf{Your Current Status} \newline Today marks the 8th day of an average menstrual cycle that lasts about once every month. You are currently experiencing what we call the 'follicular phase,' which typically occurs after menstruation and before ovulation. \par\smallskip \textbf{What Your Data Is Showing} \newline You're getting a good night's sleep, averaging around seven hours each night with your heart rate at rest being approximately 65 beats per minute (BPM). The stability in this measurement \textit{suggests} that you're managing stress well. Additionally, the oxygen levels measured by SpO2 are excellent and there's no indication of any symptoms related to these observations. \par\smallskip \textbf{Symptom Context} \newline Based on what we've observed from today's data: \par\smallskip \textbullet~Your sleep quality is good. \newline \textbullet~You haven't reported experiencing discomfort or unusual sensations today (symptoms: none). \newline \textbullet~The stability in your heart rate \textit{suggests} that you're not under significant stress at the moment. \par\smallskip Remember, this summary reflects just one day's observations and doesn't provide a complete picture of how you feel overall; it's always important to consider other factors like diet, activity level, mood changes or any discomfort you've felt throughout different parts of today. \par\smallskip If there's anything specific that's been concerning you that we haven't covered here—or if you're noticing patterns over time—please don't hesitate to share. Your health and well-being are our top priority! \\
\bottomrule
\end{tabular}
\caption{Full Phi-4-mini summaries of record MH-0094 under the three generation settings.}
\label{tab:qual_full_phi}
\end{table*}

\end{document}